%% file: main.tex
\documentclass{article}

\usepackage[margin=1in]{geometry}

\usepackage{microtype}
\usepackage{graphicx}
\usepackage{subfigure}
\usepackage{booktabs} % for professional tables

\usepackage{hyperref}

\usepackage{amsmath}
\usepackage{amssymb}
\usepackage{amsthm}
\usepackage{bbm}
\usepackage{authblk}

\usepackage{algorithm}
\usepackage{algorithmic}

\usepackage{natbib} % for citation

\input{macros} % my macro input

\usepackage{mathtools}
\mathtoolsset{showonlyrefs}
\title{Towards Practical Preferential Bayesian Optimization \\ with Skew Gaussian Processes}

\author[1,2,3]{Shion Takeno}
\author[2]{Masahiro Nomura}
\author[1]{Masayuki Karasuyama}

\affil[1]{Nagoya Institute of Technology}
\affil[2]{CyberAgent}
\affil[3]{RIKEN AIP}

\affil[ ]{\texttt{shion.takeno@riken.jp, nomura\_masahiro@cyberagent.ac.jp}}
\affil[ ]{\texttt{karasuyama@nitech.ac.jp}}
\date{}

\begin{document}
\maketitle

\begin{abstract}
We study preferential Bayesian optimization (BO) where reliable feedback is limited to pairwise comparison called duels. An important challenge in preferential BO, which uses the preferential Gaussian process (GP) model to represent flexible preference structure, is that the posterior distribution is a computationally intractable skew GP. The most widely used approach for preferential BO is Gaussian approximation, which ignores the skewness of the true posterior. Alternatively, Markov chain Monte Carlo (MCMC) based preferential BO is also proposed. In this work, we first verify the accuracy of Gaussian approximation, from which we reveal the critical problem that the predictive probability of duels can be inaccurate. This observation motivates us to improve the MCMC-based estimation for skew GP, for which we show the practical efficiency of Gibbs sampling and derive the low variance MC estimator. However, the computational time of MCMC can still be a bottleneck in practice. Towards building a more practical preferential BO, we develop a new method that achieves both high computational efficiency and low sample complexity, and then demonstrate its effectiveness through extensive numerical experiments.
\end{abstract}

% We study preferential Bayesian optimization (BO) where reliable feedback is limited to pairwise comparison called duels. An important challenge in preferential BO, which uses the preferential Gaussian process (GP) model to represent flexible preference structure, is that the posterior distribution is a computationally intractable skew GP. The most widely used approach for preferential BO is Gaussian approximation, which ignores the skewness of the true posterior. Alternatively, Markov chain Monte Carlo (MCMC) based preferential BO is also proposed. In this work, we first verify the accuracy of Gaussian approximation, from which we reveal the critical problem that the predictive probability of duels can be inaccurate. Thus, since we have to consider the skewness, we improve the MC estimator for skew GP, for which we show the low estimation variance. However, the computational time of MCMC can still be a bottleneck in practice. Therefore, we develop a new preferential BO method that achieves computational efficiency while maintaining superior optimization performance. Numerical experiments demonstrate the effectiveness of the proposed method in terms of sample complexity and computational efficiency.

\input{manuscripts/1_intro}

\input{manuscripts/2_background}
\input{manuscripts/3_skewGP}
\input{manuscripts/4_proposed}
\input{manuscripts/5_related}

\input{manuscripts/6_experiments}

\input{manuscripts/7_conclusion}

\section*{Acknoledements}
This work was supported by MEXT KAKENHI 21H03498, 22H00300, MEXT Program: Data Creation and Utilization-Type Material Research and Development Project Grant Number JPMXP1122712807, and JSPS KAKENHI Grant Number JP21J14673.

% \clearpage
\bibliography{ref}
\bibliographystyle{abbrvnat}

%%%%%%%%%%%%%%%%%%%%%%%%%%%%%%%%%%%%%%%%%%%%%%%%%%%%%%%%%%%%%%%%%%%%%%%%%%%%%%%
%%%%%%%%%%%%%%%%%%%%%%%%%%%%%%%%%%%%%%%%%%%%%%%%%%%%%%%%%%%%%%%%%%%%%%%%%%%%%%%
% APPENDIX
%%%%%%%%%%%%%%%%%%%%%%%%%%%%%%%%%%%%%%%%%%%%%%%%%%%%%%%%%%%%%%%%%%%%%%%%%%%%%%%
%%%%%%%%%%%%%%%%%%%%%%%%%%%%%%%%%%%%%%%%%%%%%%%%%%%%%%%%%%%%%%%%%%%%%%%%%%%%%%%
\newpage
\appendix
\onecolumn

\input{manuscripts/9_appendix}

\end{document}

%% file: macros.tex
\def\*#1{\boldsymbol{#1}}

\theoremstyle{plain}
\newtheorem*{rep@theorem}{\rep@title}
\newcommand{\newreptheorem}[2]{%
\newenvironment{rep#1}[1]{%
 \def\rep@title{#2 \ref{##1}}%
 \begin{rep@theorem}}%
 {\end{rep@theorem}}}

\newreptheorem{theorem}{Theorem}

\newreptheorem{lemma}{Lemma}

\newreptheorem{corollary}{Corollary}

\newreptheorem{assumption}{Assumption}

\newreptheorem{definition}{Definition}
\newtheorem{proposition}{Proposition}[section]
\newreptheorem{proposition}{Proposition}

\newcommand{\myPr}{\mathrm{Pr}}

\DeclareMathOperator*{\argmax}{arg\,max}

\newcommand{\lw}[1]{\smash{\lower2.ex\hbox{#1}}}

\newcommand{\RR}{\mathbb{R}}
\newcommand{\NN}{\mathbb{N}}

\newcommand{\EE}{\mathbb{E}}
\newcommand{\VV}{\mathbb{V}}

\newcommand{\cD}{{\cal D}}

\newcommand{\cN}{{\cal N}}

\newcommand{\cX}{{\cal X}}

\newcommand{\textbfit}[1]{\textbf{\textit{#1}}}

%% file: manuscripts/1_intro.tex
\section{Introduction}
Preferential Bayesian optimization (BO) has been an attractive approach for solving problems where reliable feedback is limited to \emph{pairwise comparison}, the so-called \emph{duels}.
This preference setting often appears in \emph{human-in-the-loop optimization} problems such as visual design optimization~\citep{koyama2020sequential} and generative melody composition~\citep{zhou2021interactive} because it is easier for humans to judge which one is better than to give an absolute rating~\citep{kahneman2013prospect}.
The system (i.e., the optimization method) in the human-in-the-loop optimization presents choices and receives preferential feedback \emph{interactively}.
To reduce the waiting time for users, the system is required to \emph{quickly} present the new options to users by learning from the observed feedback information.

An important challenge in preferential BO, which uses the preferential Gaussian process (GP) model to represent preference structure, is that the posterior distribution is computationally intractable skew GP.
The existing approaches to this difficulty are twofold:
The first approach is the Gaussian approximation, such as Laplace approximation (LA) and expectation propagation (EP)~\citep{chu2005-preference,chu2005-extensions}.
Actually, most of the preferential BO algorithms proposed so far are based on this Gaussian approximation~\citep{brochu2010-tutorial,nielsen2015-perception,gonzalez2017preferential,siivola2021-preferential,fauvel2021-efficient}, leading to good computational efficiency.
However, the predictive accuracy can be poor, as discussed in Section~\ref{sec:comparison_PGP}, because the Gaussian approximation \emph{ignores} the skewness of the exact posterior represented as the skew GP~\citep{benavoli2020skew,benavoli2021-preferential,benavoli2021-unified}.
The second approach is to directly employ the skew GP model using the Markov chain Monte Carlo (MCMC) method~\citep{benavoli2020skew,benavoli2021-preferential,benavoli2021-unified}.
%
% Although using the skew GP leads to superior optimization performance in terms of sample complexity, the MCMC requires a heavy computational time.
% %
% Reducing the computational complexity can be critical in preferential BO because, for example, the computational time directly becomes the waiting time for users in applications involving human interactions.
\citet{benavoli2020skew,benavoli2021-preferential,benavoli2021-unified} argued that the Gaussian approximation has poor estimation accuracy, and their MCMC-based estimation is effective in preferential BO.
However, the experimental evidence was insufficient to show why the Gaussian approximation degrade the preferential BO.

% In this work, we first tackle a speed-up of the MCMC estimation for skew GP.
% %
% We provide Proposition~\ref{prop:cond_SGP}, which states that the additional conditioning reduces the skew GP to GP.
% %
% From this proposition, we can analytically calculate some expectations for MC estimators, which must reduce the estimation variance, as shown in Proposition~\ref{prop:low_variance}.
% %
% Furthermore, we use the Gibbs sampling for the sampling from skew GP, which is faster than an MCMC method employed in \citep{benavoli2020skew,benavoli2021-unified} empirically.
% %
% However, since the computational time is still slower than the Gaussian approximation, exact skew GP modeling in all iterations for PBO can be unrealistic.

The aim of this study is to build truly \emph{practical} preferential BO with experimental evidence to support it.
To this end, we first evaluate the accuracy of the Gaussian approximation-based methods, which have been most used in the preferential BO literature.
Our experiments reveal that those methods fail to correctly predict the probability of duel outcomes, which is critical in the preferential BO setting.
This result implies that it is important to consider the skewness of the true posterior, and then we also improve MCMC for skew GP estimation.
Specifically, we present MC estimators that reduce the estimation variance for skew GP.
Furthermore, we empirically clarify the high computational efficiency of the Gibbs sampling compared with an MCMC method employed in \citep{benavoli2020skew,benavoli2021-unified}.
These improvements make the preferential BO algorithm with skew GP more attractive.
On the other hand, the preferential BO algorithms using MCMC can still be unrealistic in preferential BO, where the computational time may become a bottleneck in many cases.

To build practical preferential BO, we further develop a simple and computationally efficient preferential BO method that incorporates the skewness of the true posterior in a randomized manner.
% keeps the accurate prediction of probability of duels via skew GP.
The basic idea is to use a posterior additionally conditioned by a random sample from the original posterior itself, called \emph{hallucination}, by which we show that computing acquisition functions (AFs) in the proposed method is computationally efficient while reflecting the skewness of the original posterior.
Noteworthy, any powerful AFs developed so far in the standard BO literature (e.g., upper confidence bound (UCB)~\citep{Srinivas2010-Gaussian} and expected improvement (EI)~\citep{Mockus1978-Application}) can be integrated, which improves the flexibility of the proposed method.

Our contributions are summarized as follows:
\begin{enumerate}
    \item \textbfit{Rethinking existing preferential BO algorithms}: We reveal the poor accuracy of Gaussian approximation-based methods in the true posterior approximation, which implies the importance of considering \emph{skewness} in the preferential BO setting. Motivated by this observation, we improve MCMC for skew GP estimation.
    
    \item \textbfit{Building practical preferential BO algorithm}: We develop a simple and computationally efficient method, called \emph{hallucination believer}, that can reflect the skewness of the original posterior and integrate any powerful AFs developed so far in the standard BO literature.

    \item \textbfit{Extensive empirical validations}: Numerical experiments on 12 benchmark functions show that the proposed method achieves better or at least competitive performance in both terms of \emph{computational efficiency} and \emph{sample complexity} compared with Gaussian approximation-based preferential BO~\citep{gonzalez2017preferential,siivola2021-preferential,fauvel2021-efficient} and MCMC-based preferential BO~\citep{benavoli2021-preferential,benavoli2021-unified}, respectively.
\end{enumerate}

Our experimental codes are publicly available at {\small \url{ https://github.com/CyberAgentAILab/preferentialBO}}.

%% file: manuscripts/2_background.tex
\section{Background}
\subsection{Preferential Bayesian Optimization Problem}

We consider that the preferential relation is modeled by a latent function  $f: \cX \rightarrow \RR$, where $\cX \in \RR^{d}$ is the input domain.
Our goal is to maximize the latent function as
\begin{align*}
    \*x_* = \argmax_{\*x \in \cX} f(\*x),
\end{align*}
through the dueling feedback, $\*x \succ \*x^\prime$, which implies $\*x$ is preferable to $\*x^\prime$.

\subsection{Gaussian Processes for Preferential Learning}
\label{sec:skewGP}
We assume that $f$ is a sample path of GP with zero mean and some stationary kernel $k: \cX \times \cX \rightarrow \RR$.
%
% Suppose that we have multiple duels $\cD_t \coloneqq \{ \*x_{i, w} \succ \*x_{i, l} \}_{i=1}^t$, where $\*x_{i, w}$ is the winner of the duel and $\*x_{i, l}$ is the loser. % since we consider maximizing $f$ in this study.
%
Following \cite{chu2005-preference,chu2005-extensions}, we consider that the duel is determined as follows:
% \nomura{We want to note that we consider the maximization problem (in the first paragraph of this section?).}
%
% That is, given pairs of the winner and loser of \emph{duels} $\cD_t \coloneqq \{ (\*x_{i, w}, \*x_{i, l}) \}_{i=1}^t$, we consider the following relation:
% Let us define an input $\*x \in \cX \subset \RR^d$ and sample path of $\cG \cP (0, k)$, $f : \cX \rightarrow \RR$, that is determined by some kernel $k$.
% Suppose that we have multiple \emph{duels} $\cD_t \coloneqq \{ (\*x_{i, w}, \*x_{i, l}) \}_{i=1}^t$ where the duel $\*x_{i, w} \succ \*x_{i, l}$ is defined as follows:
\begin{align*}
    \*x_{i, w} \succ \*x_{i, l}
    % \Leftrightarrow
    % f(\*x_{i,  w}) - f(\*x_{i,  l}) - 2 \epsilon_t \geq 0,
    % \Leftrightarrow
    % v_t \coloneqq f(\*x_{i,  l}) - f(\*x_{i,  w}) + 2 \epsilon_t < 0,
    \Leftrightarrow
    f(\*x_{i, w}) + \epsilon_w > f(\*x_{i, l}) + \epsilon_l,
\end{align*}
where i.i.d. additive noise $\epsilon_w$ and $\epsilon_l$ follow the normal distribution $\cN(0, \sigma^2_{\rm noise})$.
%
% Since we consider in this study maximizing $f$, $\*x_{t, w}$ is the winner of the duel and $\*x_{t, l}$ is the loser.
%
This is equivalent to assuming that the preferences are obtained by which a direct observation $y = f(\*x) + \epsilon$ is bigger or smaller, where $\epsilon \sim \cN(0, \sigma^2_{\rm noise})$.
Therefore, the training data $\cD_t$ can be written as, 
\begin{align*}
    \cD_t = \{ \*x_{i, w} \succ \*x_{i, l} \}_{i=1}^t \equiv \{v_i < 0\}_{i=1}^t,
\end{align*}
where $\*x_{i, w}$ and $\*x_{i, l}$ are a winner and a loser of $i$-th duel, respectively, and $v_i \coloneqq f(\*x_{i, l}) + \epsilon_l - f(\*x_{i, w}) - \epsilon_w$.
For brevity, we denote $\{v_i < 0\}_{i=1}^t$ as $\*v_t < \*0$, where $\*v_t \coloneqq (v_1, \dots, v_t)^\top$.

The exact posterior distribution $p(f \mid \*v_t < \*0)$ is skew GP, as shown in \citep{benavoli2021-preferential,benavoli2021-unified}.
That is, for any output vector $\*f_{\rm tes} \coloneqq \bigl( f(\*x_{1, {\rm tes}}), \dots, f(\*x_{m, {\rm tes}}) \bigr)^\top$, where $m \in \NN$, we obtain the posterior distribution from Bayes' theorem as follows:
\begin{align}
    p\left(\*f_{\rm tes} \mid \*v_t < \*0 \right) 
    = \frac{\Pr\left(\*v_t < \*0 \mid \*f_{\rm tes}\right)  p\left(\*f_{\rm tes}\right)}{\Pr(\*v_t < \*0)},
    \label{eq:posterior}
\end{align}
which is the distribution called multivariate unified skew normal \citep{azzalini2013-skewnormal,benavoli2021-preferential}.

Then, we briefly derive the components of Eq.~\eqref{eq:posterior}.
Hereafter, we denote $(i,j)$-th element of matrix as $[\cdot]_{i,j}$ and both $i$-th row of matrix and $i$-th element of vector as $[\cdot]_i$.
Suppoce that $\*X \in \RR^{ (m + 2t) \times d}$ is
\begin{align*}
    \*X \coloneqq \! \bigl( \*x_{1, {\rm tes}}, \dots, \*x_{m, {\rm tes}}, \*x_{1, w}, \dots, \*x_{t, w}, \*x_{1, l}, \dots, \*x_{t, l} \bigr)^\top \!.
\end{align*}
Then, the prior for $\*f \coloneqq \bigl(f([\*X]_1), \dots, f([\*X]_{m + 2t}) \bigr)^\top \in \RR^{m + 2t}$ is,
\begin{align*}
    \*f \sim \cN(\*0, \*K),
\end{align*}
where $[\*K]_{i, j}$ is $k([\*X]_i, [\*X]_j)$.
Since $\*f_{\rm tes}$ and $\*v_t$ are linear combinations of $\*f$ and the noises $\epsilon_{w}$ and $\epsilon_{l}$, we see that,
\begin{align}
    \begin{bmatrix}
        \*f_{\rm tes} \\
        \*v_t
    \end{bmatrix}
    \sim \cN(\*0, \*\Sigma),
    \label{eq:prior_tes_v}
\end{align}
where
\begin{align}
    \*\Sigma &\coloneqq \*A \bigl(\*K + \*B \bigr) \*A^\top \in \RR^{ (m + t) \times (m + t) },
    \label{eq:prior_covariance} \\
    \*A &\coloneqq 
    \begin{bmatrix}
        \*I_{m} & \*0 & \*0 \\
        \*0 & - \*I_{t} & \*I_{t}
    \end{bmatrix} \in \RR^{ (m + t) \times (m + 2t) }, \\
    \*B &\coloneqq 
    \begin{bmatrix}
        \*0 & \*0 \\
        \*0 & \sigma^2_{\rm noise} \*I_{2t}
    \end{bmatrix} \in \RR^{(m + 2t) \times (m + 2t)}, 
\end{align}
and $\*I_{i} \in \RR^{i \times i}$ is the identity matrix.
%
% Then, from Bayes theorem, we obtain the posterior distribution,
% \begin{align}
%     p\left(\*f_{\rm tes} \mid \*v_t < \*0 \right) 
%     = \frac{\Pr\left(\*v_t < \*0 \mid \*f_{\rm tes}\right)  p\left(\*f_{\rm tes}\right)}{\Pr(\*v_t < \*0)}.
%     \label{eq:posterior}
% \end{align}
%
Thus, since the conditional distribution of multivariate normal (MVN) is MVN again \citep{Rasmussen2005-Gaussian}, $\*v_t \mid \*f_{\rm tes}$ follows MVN.
Consequently, $p\left(\*f_{\rm tes}\right)$ is a probability density function (PDF) of MVN, and $\Pr(\*v_t < \*0)$ and $\Pr\left(\*v_t < \*0 \mid \*f_{\rm tes}\right)$ are cumulative distribution functions (CDF) of MVN.
%
% Hence, Eq.~\eqref{eq:posterior} is a multivariate unified skew normal distribution \citep{azzalini2013-skewnormal,benavoli2021-preferential}.
%
% This can also be interpreted as a marginal distribution of truncated MVN (TMVN) $p(\*f_{\rm tes}, \*v | \*v < \*0)$.

% The exact posterior distribution $p(f \mid \cD_t)$ is skew GP, as shown in \cite{benavoli2021-preferential,benavoli2021-unified}.
% %
% By the definition, the training data (duels) $\cD_t$ can be rewritten as $\{v_i < 0\}_{i=1}^t$, where $v_i \coloneqq f(\*x_{i, l}) + \epsilon_l - f(\*x_{i, w}) - \epsilon_w$.
% %
% For brevity, we denote $\{v_i < 0\}_{i=1}^t$ as $\*v_t < \*0$, where $\*v_t \coloneqq (v_1, \dots, v_t)^\top$.
% %
% Then, the posterior density function for all $\*x \in \cX$ is written as,
% \begin{align*}
%     p\left(f(\*x) \mid \cD_t \right) 
%     &= p\left(f(\*x) \mid \*v_t < \*0 \right) \\
%     &= \frac{\Pr\left(\*v_t < \*0 \mid f(\*x)\right)  p\left(f(\*x)\right)}{\Pr(\*v_t < \*0)}.
% \end{align*}
% %
% Then, $\*v_t$, which we referred to as the \emph{latent truncated variable}, controls the skewness.
% %
% Since the prior for $\*v_t$ and $f(\*x)$ is multivariate normal (MVN) distribution, both $\Pr(\*v_t < \*0)$ and $\Pr\left(\*v_t < \*0 \mid f(\*x)\right)$ are cumulative distribution function (CDF) of MVN (See Appendix~\ref{app:skewGP} for the details).
% %
% Furthermore, statistics of $f(\*x) \mid \*v_t < \*0$, such as mean and variance, are computed using CDF of MVN \citep{azzalini2013-skewnormal}.
% %
% However, CDF of MVN \citep{Genz1992-numerical,Genton2017-Hierarchical} is computationally expensive.

Statistics for skew GP, such as expectation, variance, probability, and quantile, are hard to compute.
Specifically, they all need many times of computation of CDF of MVN \citep{azzalini2013-skewnormal}.
CDF of MVN \citep{Genz1992-numerical, Genton2017-Hierarchical} is computationally intractable and expensive.
Therefore, the posterior inference using CDF of MVN at each iteration for preferential BO can be unrealistic.

To approximate the statistics, \citet{benavoli2021-preferential,benavoli2021-unified} employed the MCMC, for which they showed that a sample path from skew GP $p(f | \*v_t < \*0)$ could be generated through the sampling from truncated MVN (TMVN) and MVN.
For example, they approximate the posterior mean $\EE_{f | \*v_t < \*0} \bigl[ [\*f_{\rm tes}]_i \bigr]$ as,
\begin{align}
    \EE_{f | \*v_t < \*0} \bigl[ [\*f_{\rm tes}]_i \bigr] 
    \approx \frac{1}{M} \sum_{j=1}^M f_j (\*x_{i, {\rm tes}}),
    \label{eq:samplepath_estimation}
\end{align}
where $M$ is the number of MC samples and $f_j$ is a sample path from the posterior $f | \*v_t < \*0$.
%
% However, generating a sufficient number of MC samples is still time-consuming.
%
Another popular approximation is Gaussian approximation using LA and EP \citep{chu2005-preference,chu2005-extensions}.
LA \citep{Mackay1996-bayesian} approximates the posterior by Gaussian distribution so that the mean is equal to the mode of the true posterior.
EP \citep{Minka2001-Minka} is an iterative moment-matching method, which aims to adjust the first and second moments.

\citet{benavoli2020skew} revealed that the true posterior of GP classification (GPC) with probit likelihood is also a skew GP.
Therefore, in principle, the approximate inferences for GPC can apply to preferential GP.
%
% The survey of GP classification \citep{Nickisch2008-approximations,Kuss2005-assessing} recommended EP, whose prediction is very accurate for GP classification.
Actually, the survey of GPC \citep{Nickisch2008-approximations,Kuss2005-assessing} argues that EP exhibits excellent prediction performance \emph{in the GPC setting}.
However, the validity of using EP in the preferential BO has not been verified in detail so far.
Then, we will demonstrate in this study the predictive probability of the duel using EP can be inaccurate in Section~\ref{sec:comparison_PGP}.

%% file: manuscripts/3_skewGP.tex
\section{Rethinking Estimation of Preferential GP}
\label{sec:preferential GP_learning}
First, we investigate the limitation of the Gaussian approximation, which is the inaccurate predictive probability.
Second, we tackle the improvement of MCMC-based estimation for skew GP.
We show the practical effectiveness of Gibbs sampling for skew GP.
Furthermore, we newly derive MC estimators, showing that the estimation variance is lower than the estimators used in \citep{benavoli2021-preferential,benavoli2021-unified}.
%
% Although we perform experiments on the Holder table, Ackley, and Hartmann6 functions, we show the results on the Ackley function in this section, and others are shown in Appendix~\ref{app:additional_exp}.

\begin{figure}[t]
    \centering
    \includegraphics[width=0.5\linewidth]{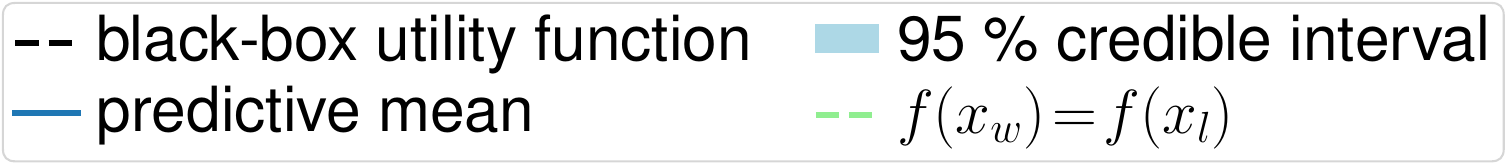}
    \hfill
    \includegraphics[width=0.6\linewidth]{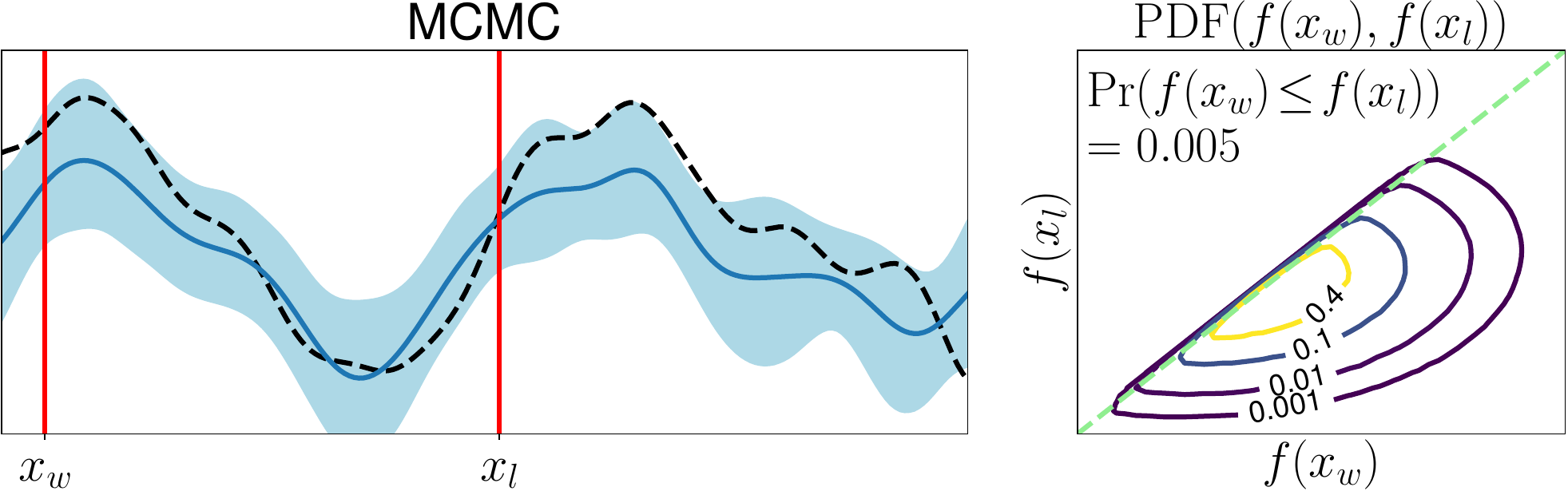}
    \hfill
    \includegraphics[width=0.6\linewidth]{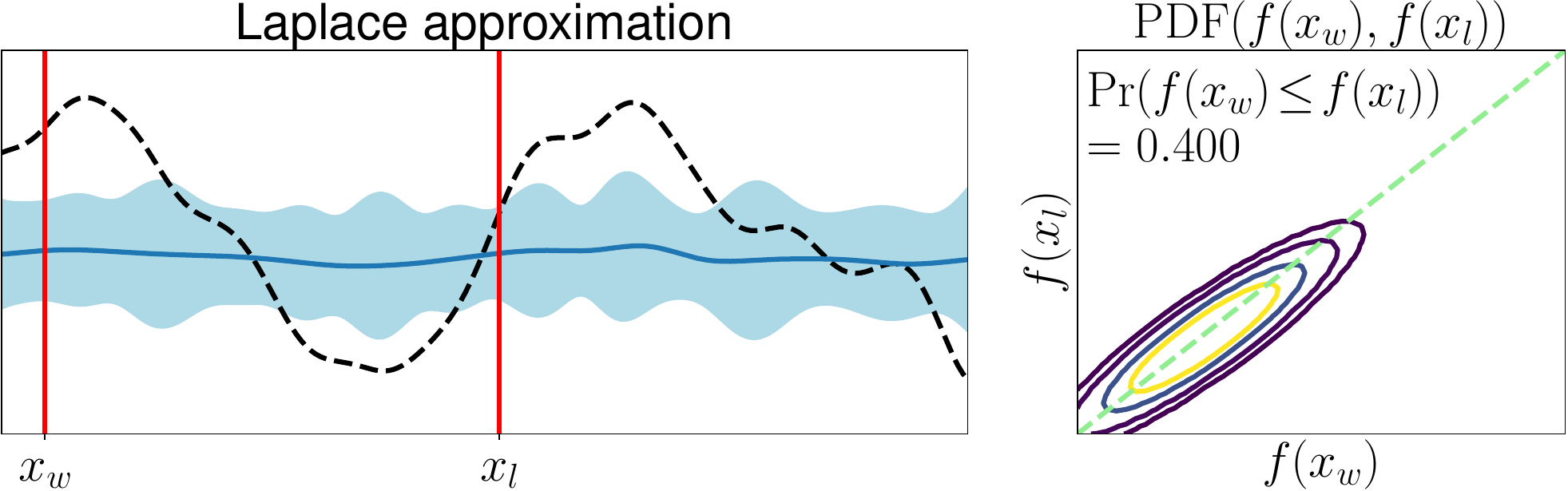}
    \hfill
    \includegraphics[width=0.6\linewidth]{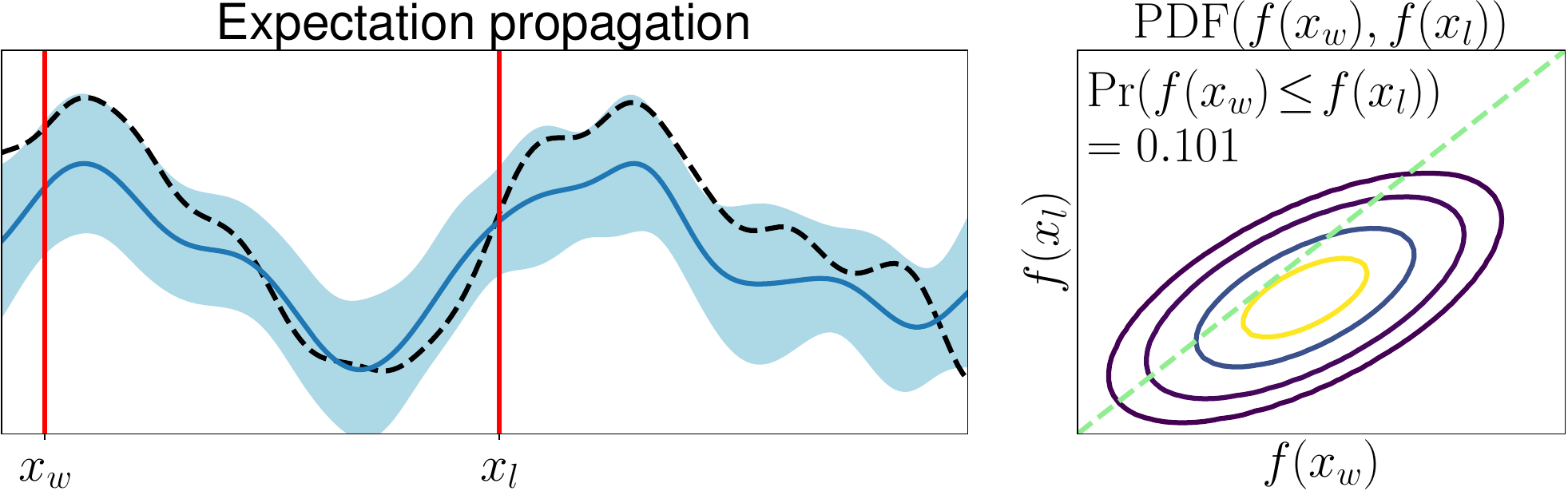}
    \caption{
        An illustrative example of preferential GP models fitted to 50 uniformly random duels, including $\*x_w \succ \*x_l$, from a sample path from GP.
        We use the RBF kernel, which is used to generate the sample path, and set $\sigma^2_{\rm noise} = 10^{-4}$.
        The top, middle, and bottom plots show the predictions of MCMC, LA, and EP, respectively.
        The left plot shows the mean and 95\% credible interval of $f | \*v_t < \*0$.
        The right plots show the estimator of $\Pr_t\bigl( f(\*x_1) \leq f(\*x_2) \bigr)$ and PDF $p(f(\*x_1), f(\*x_2) | \*v_t < \*0)$, for which an approximation by Eq.~\eqref{eq:posterior} using CDF of MVN is used in the top right plot.
        }
        \label{fig:examples_preferential GP}
\end{figure}

\begin{figure}[!t]
    \centering
    \includegraphics[width=0.2\linewidth]{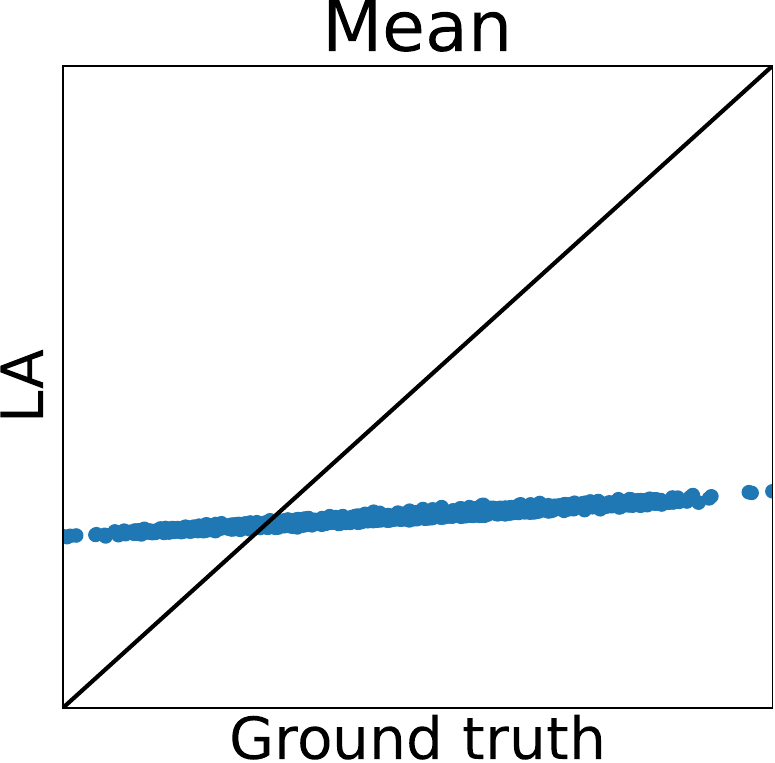}
    \includegraphics[width=0.2\linewidth]{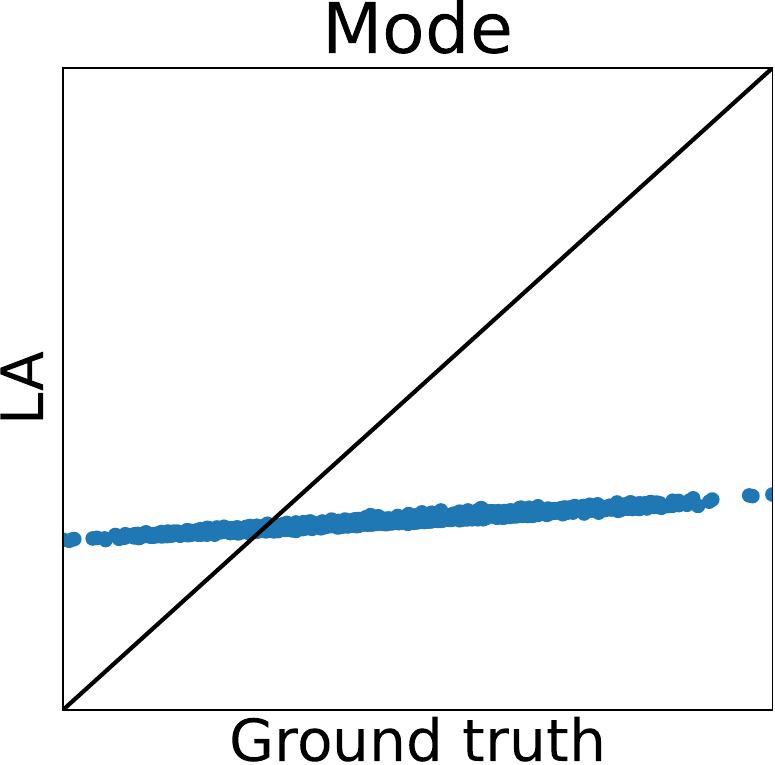}
    \includegraphics[width=0.2\linewidth]{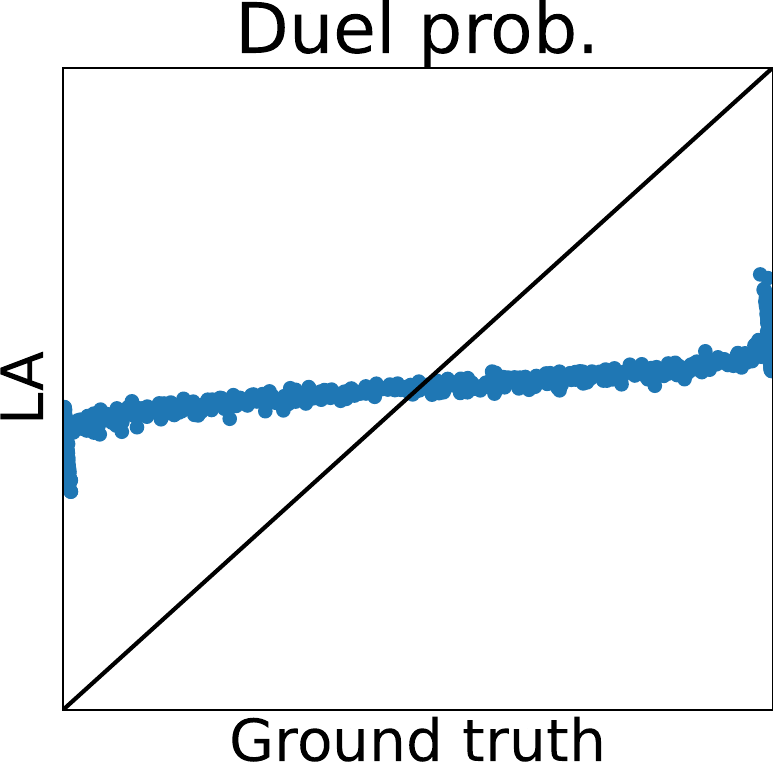}
    \subfigure[Truth vs. prediction plots for LA and EP.]{
        \centering
        \includegraphics[width=0.2\linewidth]{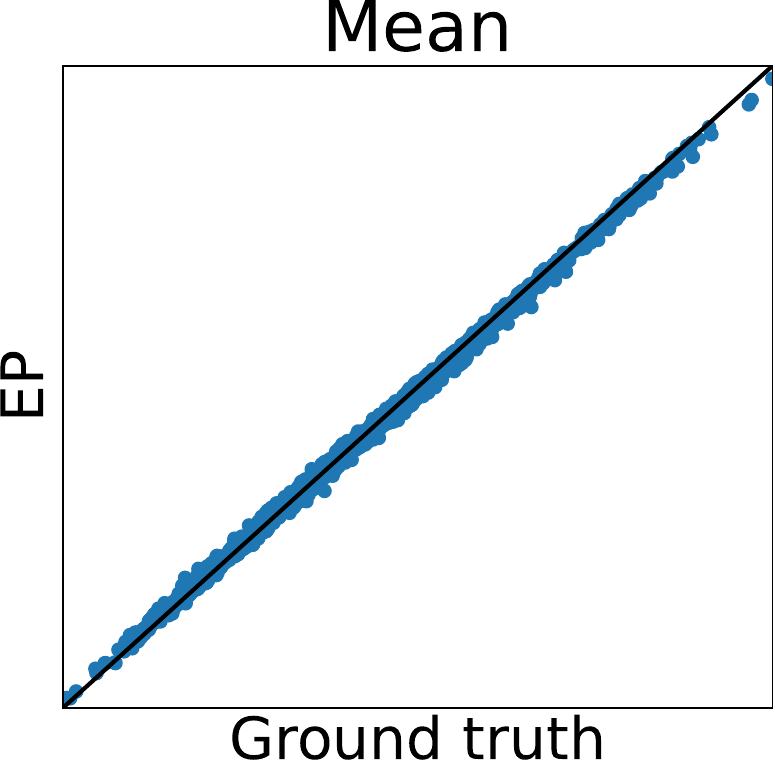}
        \includegraphics[width=0.2\linewidth]{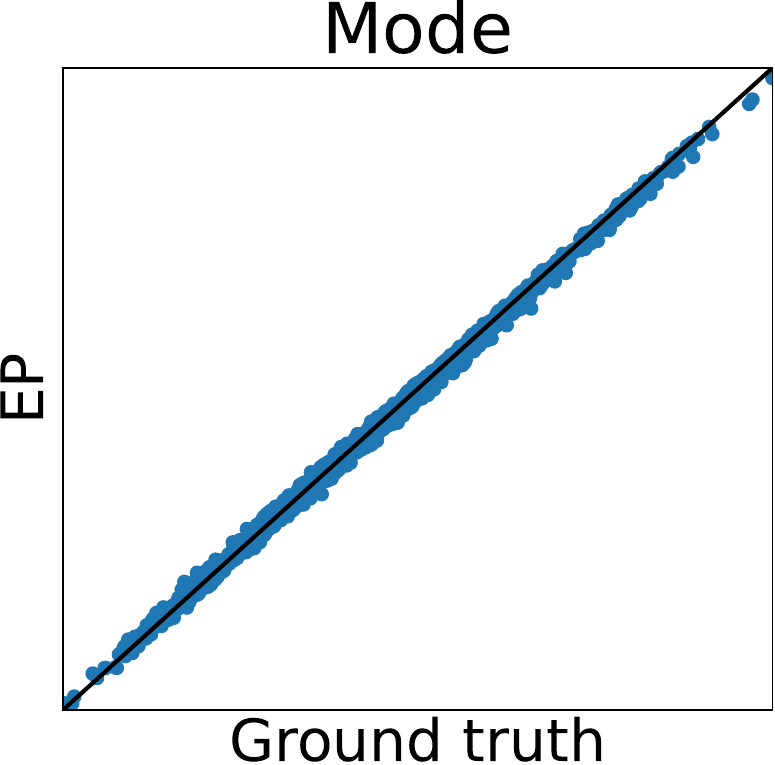}
        \includegraphics[width=0.2\linewidth]{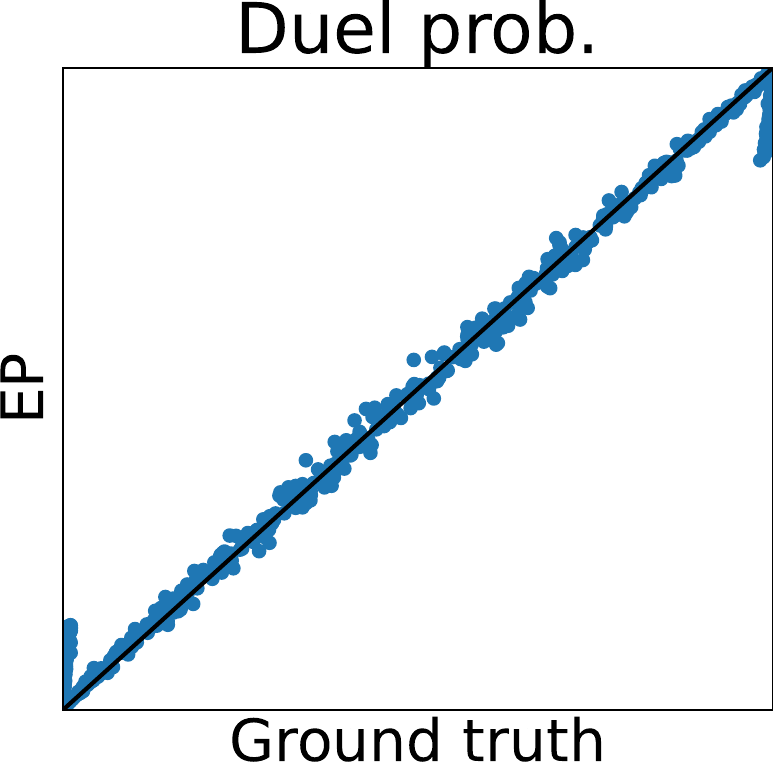}
    }
    
    \subfigure[RMSE of proposed MC estimator and full MC estimator \citep{benavoli2021-preferential} against ground truth.]{
        \includegraphics[width=0.215\linewidth]{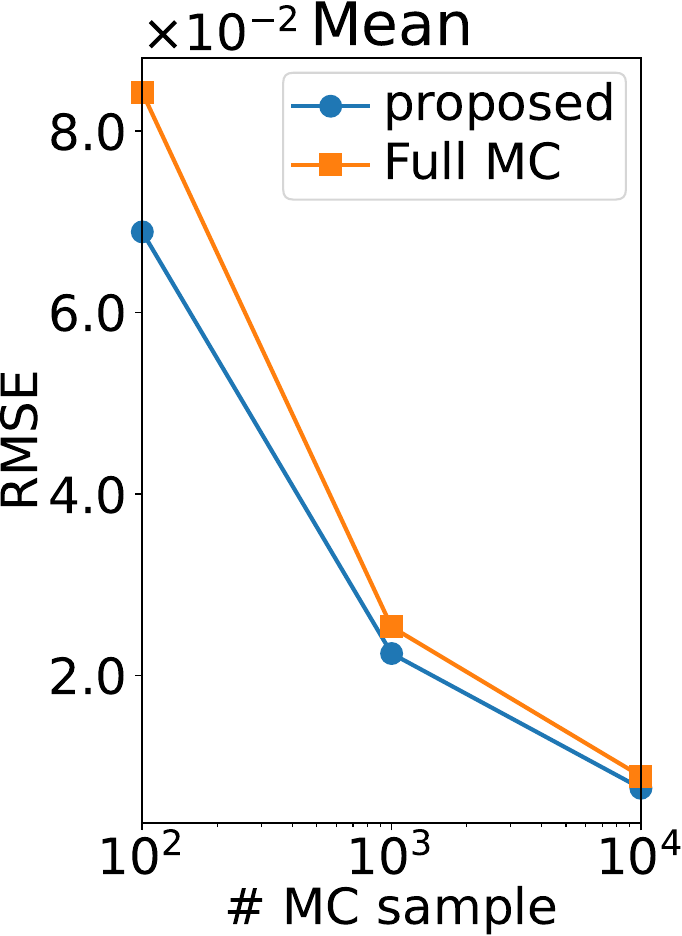}
        \includegraphics[width=0.2\linewidth]{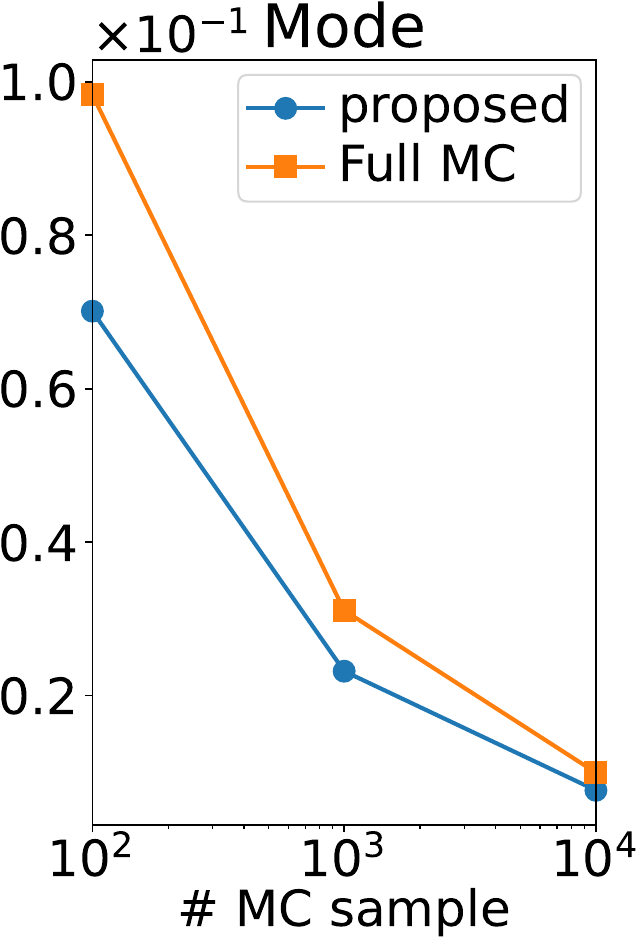}
        \includegraphics[width=0.2\linewidth]{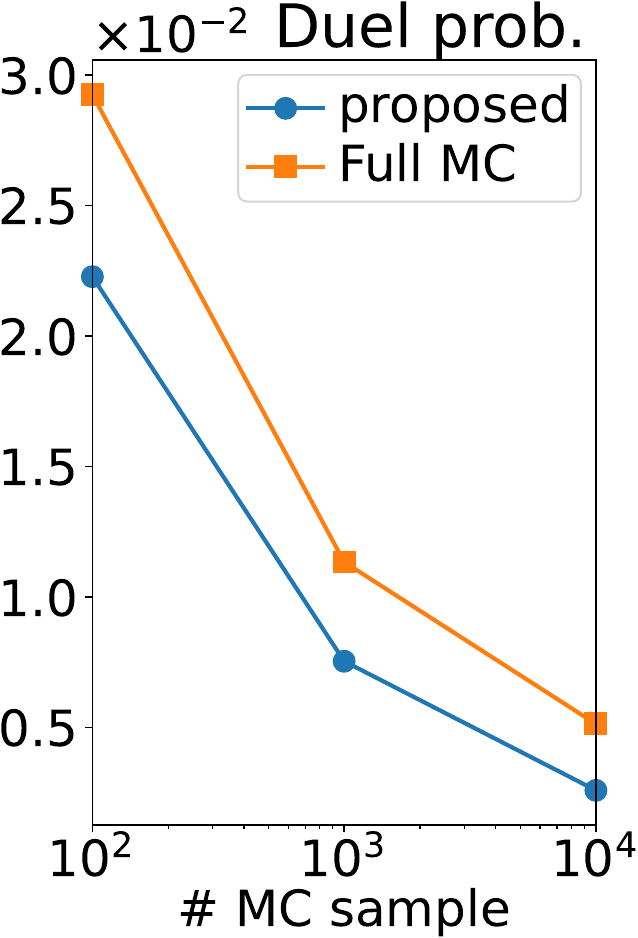}
    }
    
    \caption{
        Experimental results in the Ackley function ($d=4$).
        We evaluate the estimators of the mean, mode, and duel probabilities $\myPr_t\bigl( f(\*x) \succ f(\*x^\prime) \bigr)$.
        Other details are shown in Appendix~\ref{app:exp_settings}.
        A similar tendency was confirmed in the Holder table ($d=2$) and Hartmann6 ($d=6$) functions, which are shown in Appendix~\ref{app:additional_exp}.
        }
    \label{fig:pgp_ackley}
\end{figure}

% We fitted LA-, EP-, and MCMC-based preferential GP with RBF kernel to 50 uniformly random duels in 10 random trials.
% %
% We set $\sigma^2_{\rm noise} = 10^{-4}$, and the hyperparameters of RBF kernel are determined by marginal likelihood maximization \citep{Rasmussen2005-Gaussian}, whose details are shown in Appendix~\ref{app:exp_settings}.
% %
% We evaluate the estimators of the mean, mode, and the duel probabilities $\myPr_t\bigl( f(\*x) \succ f(\*x^\prime) \bigr)$ at 100 uniformly random inputs and inputs included in the training duels.
% %
% The MCMC uses 10000 MC samples, 1000 burn-in, and 10 thinning.

%%%%%%%%%%%%%%%%%%%%%%%%%%%%%%%%%%%%%%%%%%%%%%%%%%%%%%%%%%%%%%%%%%%%%%%%%%%%%%%%%%%%%%%%%%%%%%%%%%%%%%%%%%%%%%%%%%
\subsection{Limitation of Gaussian Approximation-Based Preferential GP}
\label{sec:comparison_PGP}

To clarify the inaccuracy of Gaussian approximation, Figure~\ref{fig:examples_preferential GP} shows an illustrative example of the estimation of skew GP via MCMC, LA, and EP.
%
% Uniformly random 20 duels are provided, and $\sigma^2_{\rm noise} = 10^{-2}$.
%
The MCMC-based estimation uses many (10000) MC samples with 1000 burn-in and 10 thinning generated by Gibbs sampling, whose details are discussed in Section~\ref{sec:Gibbs_sampling} and Appendix~\ref{app:detail_gibbs_sampling}.
Thus, we can expect that the MCMC is sufficiently precise and treat it as the ground truth.
Then, we will confirm the differences between the ground truth (MCMC), and LA and EP, respectively.
% Note that in the top right plot, PDF is approximated by Eq.~\eqref{eq:posterior} using CDF of MVN.

First, we can see that the LA is inaccurate since the mode can be very far away from the mean, particularly when $\sigma^2_{\rm noise}$ is small.
A similar problem for GPC is discussed in \citep[][Section~12.2.1]{Nickisch2008-approximations}.
Therefore, although LA is very fast, LA-based preferential BO will fail.

Second, EP is very accurate for estimating the mean and the credible interval of $f | \*v_t < \*0$, which matches the experimental observation in the literature of GPC \citep{Nickisch2008-approximations,Kuss2005-assessing}.
% \footnote{This contradicts the description that EP overestimates the variance, in \citep[][pp.16]{benavoli2021-unified}. However, although \citet[][pp. 16]{benavoli2021-unified} wrote that \citet{Nickisch2008-approximations,Kuss2005-assessing} described the same claim, we could not find such a description. Furthermore, the EP is represented as ``good approximation'' in \citep[][Table~1]{Nickisch2008-approximations}.}.
%
However, in the right plot, $\Pr_t (f(\*x_w) \leq f(\*x_l))$ is overestimated although $\*x_w \succ \*x_l$ is already obtained.
This implies the problem of EP that predictions for the statistics involving the joint distribution of several outputs (e.g., $f(\*x_w)$ and $f(\*x_l)$) can be inaccurate.
Particularly, since some AFs for preferential BO \citep{gonzalez2017preferential,nielsen2015-perception,fauvel2021-efficient} need to evaluate the joint distribution of $f(\*x^{(1)}_t)$ and $f(\*x^{(2)}_t)$, EP can degrade the performance of such AFs.
This is a crucial problem but has not been discussed in the literature on preferential GP \citep{chu2005-extensions,benavoli2021-preferential,benavoli2021-unified} and GPC \citep{Kuss2005-assessing,Nickisch2008-approximations}.
%
% This problem of the predictive probability for duels is not discussed in \citep{chu2005-extensions,benavoli2020skew,benavoli2021-unified}
% %
% In addition, since this type of probability, which involves two outputs (e.g., $f(\*x_w)$ and $f(\*x_l)$), is not of interest in GPC, this problem also has not been considered in the literature of CBO.
% %
% On the other hand, for preferential learning, this quantity is crucial.
% %
% Furthermore, since some AFs for preferential BO need to evaluate the correlation, EP can degrade the performance of such AFs.
%
% Note that although the predictive probability of EP for GPC is described as ``very accurate'' in \citep[][Section~12.2.2]{Nickisch2008-approximations}, there is an important difference between preferential GP and GPC.
% %
% The predictive probability in GPC is defined by the distribution of an output $f(\*x)$.
% %
% In contrast, for preferential GP, we are interested in the probability of the duels, which involves the distribution of two outputs (e.g., $f(\*x_w)$ and $f(\*x_l)$).
% %
% We conclude that the prediction of EP for the statistics involving the correlation between two outputs can be unreliable.
% %
% Since some AFs for preferential BO need to evaluate the correlation, EP can degrade the performance of such AFs.

It should be noted that \citet{benavoli2021-unified} already provided the experimental comparison between EP and skew GP.
However, we argue that their setting is not common in the following aspects:
For example, \citep[][Figure~6 and 7]{benavoli2021-unified} considers that all the inputs in a certain interval lose, and all other inputs in another interval win.
Although the Gaussian approximation becomes inaccurate due to the heavy skewness in this setting, these biased training duels are unrealistic.
In contrast, Figure~\ref{fig:examples_preferential GP}, in which the usual uniformly random duels are used, considers a more common setting.

Figure~\ref{fig:pgp_ackley}~(a) further shows the truth (MCMC) and predictions of LA and EP plots using the Ackley function.
We can again confirm that LA underestimates all the statistics, and the mean and mode estimators of EP are very accurate.
However, for the probability of duels, EP underestimates the probabilities around 0 or 1.
Importantly, we can see that the estimator does not retain the relative relationship of the ground truth.
Thus, EP-based AF can select the different inputs from the inputs selected by the ground truth.
Therefore, Figure~\ref{fig:pgp_ackley}~(a) also suggests that EP-based preferential BO can deteriorate.

%%%%%%%%%%%%%%%%%%%%%%%%%%%%%%%%%%%%%%%%%%%%%%%%%%%%%%%%%%%%%%%%%%%%%%%%%%%%%%%%%%%%%%%%%%%%%%%%%%%%%%%%%%%%%%%%%%

\subsection{Improving MCMC-based Preferential GP}
% Due to the above limitations of the Gaussian approximation, we need to evaluate skew GP without Gaussian approximation.
% %
% Therefore, we work to improve the MCMC for skew GP.
The above results motivated us to improve the MCMC-based estimation of the skew GP.
In this section, we show the practical effectiveness of Gibbs sampling and derive the low variance MC estimators.

\begin{figure}[!t]
    \centering
    \includegraphics[width=0.7\linewidth]{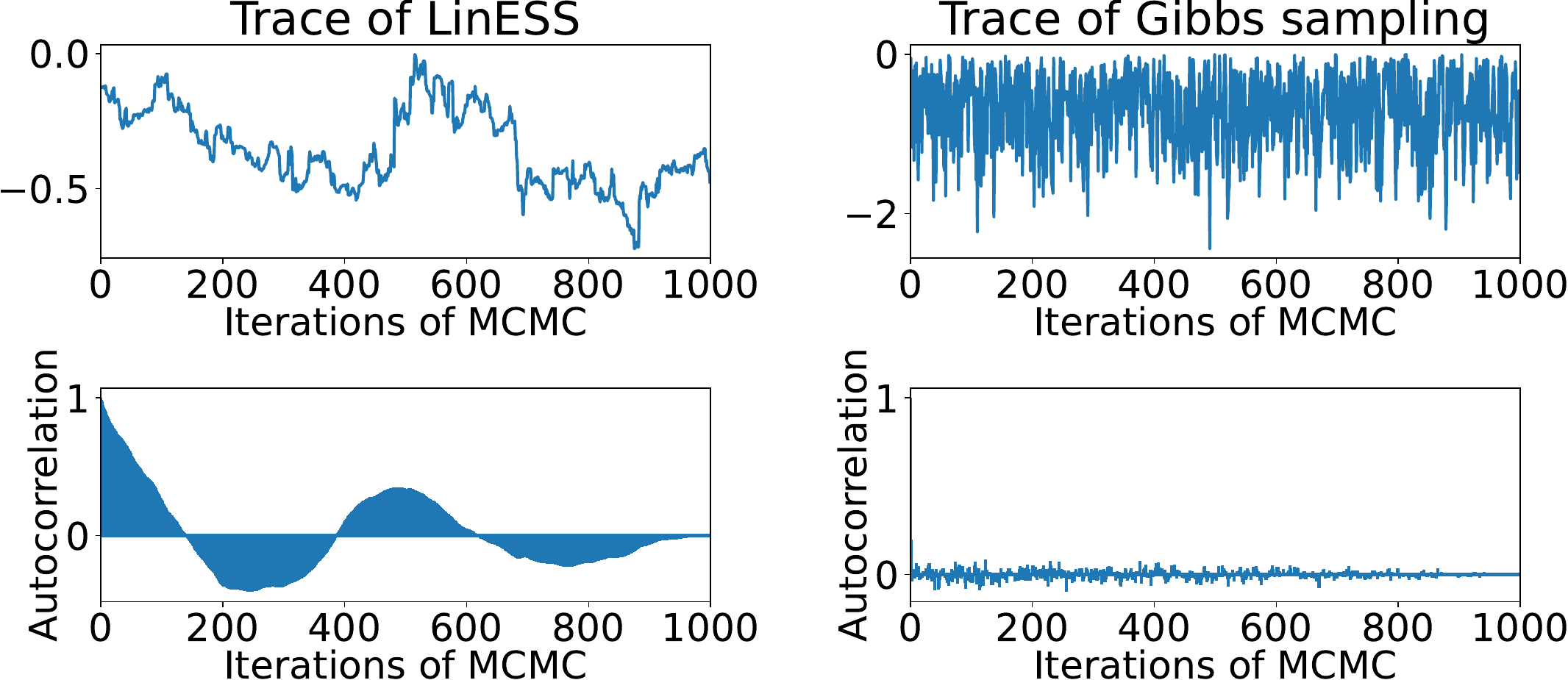}
    \caption{Trace and autocorrelation plots of LinESS and Gibbs sampling for first 1000 samples used in the experiments in Figure~\ref{fig:pgp_ackley}.}
    \label{fig:trace_ackley}
\end{figure}

\begin{table}[!t]
    \caption{The mean and standard deviation of computational time of MCMC for Figure~\ref{fig:trace_ackley} with 10 random trials.}
    \label{tab:computation_time_MCMC_ackley}
    \centering
    \begin{tabular} {c|cccc} 
        & LinESS & Gibbs sampling \\ \hline 
        Time (sec) & $1.61 \pm 0.01$ & $0.54 \pm 0.01$
    \end{tabular}
\end{table}

%%%%%%%%%%%%%%%%%%%%%%%%%%%%%%%%%%%%%%%%%%%%%%%%%%%%%%%%%%%%%%%%%%%%%%%%%%%%%%%%%%%%%%%%%%%%%%%%%%%%%%%%%%%%%%%%%%
\subsubsection{Gibbs Sampling for TMVN}

\label{sec:Gibbs_sampling}
In this study, for MCMC of skew GP that requires sampling from TMVN, we argue that the Gibbs sampling should be employed rather than linear elliptical slice sampling (LinESS) \citep{gessner2020-integrals}, which is used in the existing studies~\citep{benavoli2021-preferential,benavoli2021-unified}.
% , which is required to sample from TMVN.
% The key fact is that estimation of skew GP requires the sampling of $\*v_t$ from TMVN.
%
% \citet{benavoli2021-preferential,benavoli2021-unified} used linear elliptical slice sampling (LinESS) \cite{gessner2020-integrals}, the variant of elliptical slice sampling \cite{murray2010elliptical}.
% %
% Since LinESS is a rejection-free sampling method, LinESS is expected to be effective when the number of truncations is larger than $t$, as discussed in \citep[][Section~2]{gessner2020-integrals}.
% %
% On the other hand, in the estimation of skew GP, the number of truncations and the dimension is always equal.
%
% Therefore, the effectiveness of LinESS is limited in our experiments, as we will show in Section~\ref{sec:exp}.
%
Gibbs sampling is a standard approach for the sampling from TMVN \citep{adler2008-efficient,Breslaw1994-random,geweke1991efficient,kotecha1999gibbs,li2015efficient,robert1995-simulation}.
In Gibbs sampling, we need to sample univariate truncated normal, for which we used the efficient rejection sampling \citep[][Section 2.1]{li2015efficient}, which uses several proposal distributions depending on the truncation.
Detailed procedure and pseudo-code of Gibbs sampling are shown in Appendix~\ref{app:detail_gibbs_sampling}.

Figure~\ref{fig:trace_ackley} shows the trace and autocorrelation plots of LinESS and Gibbs sampling, where we used the implementation by \citep{benavoli2021-preferential,benavoli2021-unified} for LinESS ({\small \url{https://github.com/benavoli/SkewGP}}).
The mixing time of Gibbs sampling is very fast compared with LinESS.
Furthermore, as shown in Table~\ref{tab:computation_time_MCMC_ackley}, Gibbs sampling is roughly 3x faster than LinESS.
Consequently, our experimental results suggest that Gibbs sampling is more suitable than LinESS for preferential GP.

Note that LinESS is expected to be effective when the number of truncations is much larger than the dimension of MVN, given that LinESS is a rejection-free sampling method.
In contrast, when the number of truncations is huge, a rejection sampling-based Gibbs sampling suffers a low acceptance rate.
On the other hand, in this case for $\*v_t \mid \*v_t < \*0$, both dimensions are $t$.
We conjecture that this is the reason why Gibbs sampling is efficient compared to LinESS in our experiments.

%%%%%%%%%%%%%%%%%%%%%%%%%%%%%%%%%%%%%%%%%%%%%%%%%%%%%%%%%%%%%%%%%%%%%%%%%%%%%%%%%%%%%%%%%%%%%%%%%%%%%%%%%%%%%%%%%%
\subsubsection{MC Estimation for Posterior Statistics}
\label{sec:procedure_inference}

% We derive the MC estimator for skew GP, whose estimation variance is must smaller than that of \citep{benavoli2021-preferential,benavoli2021-unified}, e.g., Eq.~\eqref{eq:samplepath_estimation}.
%
% Here, let us consider the case of the posterior mean for brevity.
%
To derive the low variance MC estimator, we use the following fact:
\begin{proposition}
    The exact posterior distribution additionally conditioned by $\*v_t$ is a GP.
    That is, for an arbitrarily output vector $\*f_{\rm tes}$, we obtain
    \begin{align}
        \*f_{\rm tes} \mid \*v_t, \*v_t < \*0
        &\stackrel{\rm{d}}{=} \*f_{\rm tes} \mid \*v_t 
        \sim \cN(\*\mu_{{\rm tes} | \*v}, \*\Sigma_{{\rm tes} | \*v}), \\
        \*\mu_{{\rm tes} | \*v} &\coloneqq \*\Sigma_{{\rm tes}, \*v} \*\Sigma_{\*v, \*v}^{-1} \*v_t, \label{eq:mu_given_v} \\
        \*\Sigma_{{\rm tes} | \*v} &\coloneqq \*\Sigma_{{\rm tes}, {\rm tes}} - \*\Sigma_{{\rm tes}, \*v} \*\Sigma_{\*v, \*v}^{-1} \*\Sigma_{{\rm tes}, \*v}^\top,
    \end{align}
    where $\stackrel{\rm{d}}{=}$ implies the equality in distribution, and we denote the sub-matrices of Eq.~\eqref{eq:prior_covariance} as follows: 
    \begin{align*}
        \*\Sigma
        \eqqcolon
        \begin{bmatrix}
            \*\Sigma_{{\rm tes}, {\rm tes}} & \*\Sigma_{{\rm tes}, \*v} \\
            \*\Sigma_{{\rm tes}, \*v}^\top & \*\Sigma_{\*v, \*v}
        \end{bmatrix},
    \end{align*}
    where $\*\Sigma_{{\rm tes}, {\rm tes}} \in \RR^{m \times m}, \*\Sigma_{{\rm tes}, \*v} \in \RR^{m \times t}, \*\Sigma_{\*v, \*v} \in \RR^{t \times t}$.
    \label{prop:cond_SGP}
\end{proposition}
See Appendix~\ref{app:proof} for the proof.

One of the important consequences of Proposition~\ref{prop:cond_SGP} is that an expectation for the posterior inference can be partially calculated by an analytical form.
For example, the posterior mean $\EE_{f | \*v_t < \*0 } [\*f_{\rm tes} ]$ can be derived as,
\begin{align}
    \EE_{f | \*v_t < \*0 } [\*f_{\rm tes}]
    &= \EE_t \bigl[ \EE_{f | \*v_t, \*v_t < \*0} [\*f_{\rm tes} ] \bigr] \\
    % &= \EE_t \bigl[ \EE_{f | \*v_t} [\*f_{\rm tes} ] \bigr] && (\because \text{Prop.~\ref{prop:cond_SGP}})\\
    &= \EE_t [ \*\mu_{{\rm tes} | \*v} ] && (\because \text{Prop.~\ref{prop:cond_SGP}}) \\
    &= \*\Sigma_{{\rm tes}, \*v} \*\Sigma_{\*v, \*v}^{-1} \EE_t [\*v_t],
    \label{eq:mean_estimator}
\end{align}
where we denote $\EE_t [\cdot] \coloneqq \EE_{\*v_t | \*v_t < \*0} [\cdot]$.
Note that, in the first and last equalities, we use the tower property of the expectation and Eq.~\eqref{eq:mu_given_v}, the definition of $\*\mu_{{\rm tes} | \*v}$, respectively.
Other statistics, such as variance and probability, can be estimated via the following derivation:
\begin{align}
    &\VV_{f | \*v_t < \*0} [\*f_{\rm tes} ] = \VV_t [\*\mu_{{\rm tes} | \*v}] + \*\Sigma_{{\rm tes} | \*v}, \\
    &\myPr \bigl( [\*f_{\rm tes}]_i  \leq c | \*v_t < \*0  \bigr) 
    = \EE_t \left[ \Phi \left( 
    \frac{c - [\*\mu_{{\rm tes} | \*v}]_i}{ \sqrt{[\*\Sigma_{{\rm tes} | \*v}]_{i,i} }}
    \right) \right], \label{eq:CDF_estimator} \\
    % &\myPr_t \bigl( \*x_{j, {\rm tes}}  \succ \*x_{i, {\rm tes}} \bigr) 
    &\myPr_t \bigl( [\*f_{\rm tes}]_i  \leq [\*f_{\rm tes}]_j | \*v_t < \*0  \bigr) \\ 
    &= \EE_t \left[ \Phi \left( 
    \frac{[\*\mu_{{\rm tes} | \*v}]_j - [\*\mu_{{\rm tes} | \*v}]_i}{ \sqrt{[\*\Sigma_{{\rm tes} | \*v}]_{i,i} + [\*\Sigma_{{\rm tes} | \*v}]_{j,j} - 2 [\*\Sigma_{{\rm tes} | \*v}]_{i,j} }}
    \right) \right],
\end{align}
where $\VV$ means the (co)variance, and we denote $\VV_t [\cdot] \coloneqq \VV_{\*v_t | \*v_t < \*0} [\cdot]$, and $\Phi$ is the CDF of the standard normal distribution.
See Appendix~\ref{app:detailed_estimators} for detailed derivation.
As a result, we only need to apply MC estimation using the samples of $\*v_t$ to $\VV_t$ or $\EE_t$.
Note that after we obtain the samples of $\*v_t | \*v_t < \*0$ once, we can reuse it for all the estimations.

The remaining important statistic is the quantile of $[\*f_{\rm tes}]_i | \*v_t < \*0$, which is useful to compute the credible interval and UCB-based AFs.
Although this can be derived as an inverse function of Eq.~\eqref{eq:CDF_estimator}, the analytical form of this inverse function is unknown.
Thus, we compute the quantile by applying a binary search to the MC estimator of Eq.~\eqref{eq:CDF_estimator}.
See Appendix~\ref{app:detailed_estimators} for details.

The error of the MC estimator is in proportion to the variance of the MC estimator.
Then, our MC estimators are justified in terms of the error of the MC estimator as follows:
%
% The following lemma implies that the variances of our MC estimators are must smaller than those of \citep{benavoli2021-preferential}:
\begin{proposition}
    The variances of our MC estimators are smaller than those of \citep{benavoli2021-preferential}, such as Eq.~\eqref{eq:samplepath_estimation}.
    For example, the following inequality holds:
    \begin{align}
        \VV_t \bigl[  [\*\mu_{{\rm tes} | \*v}]_{i} \bigr] \leq \VV_{f | \*v_t < \*0} \bigl[ [\*f_{\rm tes}]_{i} \bigr],
    \end{align}
    for all $i \in \{1, \dots, m\}$.
    \label{prop:low_variance}
\end{proposition}
This proposition is a direct consequence of the law of total variance.
See Appendix~\ref{app:proof_variance} for details.
Thus, in principle, our MC estimators are always superior to those of \citep{benavoli2021-preferential,benavoli2021-unified}.

Figure~\ref{fig:pgp_ackley}~(b) shows the root mean squared error (RMSE) of our MC estimators and those of \citep{benavoli2021-preferential,benavoli2021-unified}, which we refer to as a full MC estimator.
Note that the RMSE is computed using the ground truth.
Thus, if the ground truth is sufficiently accurate (equal to the true expectation), the RMSE can be interpreted as the standard deviation estimator of the MC estimator.
We can observe that the RMSE of our estimator is smaller than that of the full MC estimator in all the statistics and all the number of MC samples.
Therefore, the efficiency of our MC estimator was verified also experimentally.

%% file: manuscripts/4_proposed.tex
\section{Preferential BO with Hallucination Believer}
\label{sec:proposed}

Although we tackled the speed-up of MCMC-based estimation, its computational time can still be unrealistic in preferential BO.
Furthermore, Gaussian approximation often suffers poor prediction performance ignoring the skewness of the posterior.
In this work, we propose a computationally efficient preferential BO method while considering the skewness in a randomized manner.

\subsection{Hallucination Believer}

We proposed a randomized preferential BO method, called hallucination believer (HB), which uses the AFs for the standard BO calculated from GP $f | \*v_t = \tilde{\*v}_t$, where $\tilde{\*v}_t \sim p(\*v_t | \*v_t < \*0)$ is an i.i.d. sample and called \emph{hallucination}.

% There are two key points: the conditioning on $\*v_t$ and using the hallucination $\tilde{\*v}_t$ for conditioning.
% %
% First, as shown in Proposition~\ref{prop:cond_SGP}, the conditioning on $\*v_t$ reduces the posterior to the GP $f | \*v_t = \tilde{\*v}_t$, which enables us to use any computationally efficient and powerful AF for the standard BO.
% %
% Second, by using hallucination $\tilde{\*v}_t$ from the true posterior, we aim to incorporate the skewness of the posterior in a randomized manner.
% %
% This is because, if we condition as $ \*v_t = \*c$, where $\*c$ is some constant vector such as the posterior mean, the proposed method ignores skewness, and the optimization performance will deteriorate.
% %
% Consequently, we need only one sample $\tilde{\*v}_t$ from TMVN and the usual GP-based AF calculations.
% %
% Therefore, the computational complexity is significantly cheaper than MCMC-based AFs \citep{benavoli2021-preferential}.
% %
% We will show the effectiveness of this hallucination-based randomization via extensive experiments in Section~\ref{sec:exp}.

The HB method is motivated by Proposition~\ref{prop:cond_SGP}, which illustrates that the conditioning on $\*v_t$ reduces the posterior to the GP $f | \*v_t$.
As a result, we can employ any computationally efficient and powerful AF for the standard BO by utilizing the GP $f | \*v_t$.
Then, selecting the variables for the conditioning on $\*v_t$ is important.
At first glance, natural candidates may be the mean $\EE[\*v_t]$ or mode.
However, the computation of these statistics can already be  computationally expensive, as discussed in Section~\ref{sec:skewGP}.
Furthermore, since TMVN $p(\*v_t | \*v_t < \*0)$ can be highly skewed, the conditioning by mean or mode can degrade the performance as with LA and EP.
Hence, we consider using the hallucination $\tilde{\*v}_t$.
One sampling of $\tilde{\*v}_t \sim p(\*v_t | \*v_t < \*0)$ can be performed efficiently by Gibbs sampling as shown in Section~\ref{sec:Gibbs_sampling}.
In addition, the calculation of AF itself is also cheaper than that of an MCMC-based preferential BO method~\citep{benavoli2021-preferential}, which necessitates averaging over the many MCMC samples.
Lastly, we aim to incorporate the skewness of the posterior in a randomized manner by using the hallucination $\tilde{\*v}_t$.

\paragraph{Intuitive Rationale behind HB:}
We believe that it is better to consider the prediction and the optimization separately: 
For example, Thompson sampling (TS) selects the next duels using a random sample path. 
An essential requirement in this context is that the sample path should reflect the uncertainty of the true posterior rather than pursuing the accuracy as a single prediction.
In the case of the HB method, the posterior for the AF calculation is constructed based on the conditioning on a random sample $\tilde{\*v}_t$.
This sampling process also incorporates the uncertainty of the true posterior into the subsequent AF calculation, and it does not necessarily aim to recover the exact posterior at that iteration.
It is worth noting that the sampling in the HB method is only for $\tilde{\*v}_t$, in contrast to TS which samples the entire objective function.

\begin{algorithm}[t]
    \caption{Hallucination believer for preferential BO}\label{alg:HB}
    \begin{algorithmic}[1]
        \REQUIRE $\cD_0 = \{\*x_{0,w}, \*x_{0,l} \}$, $\cX$
        \FOR{$t = 1, \dots$}
            \STATE $\*x_{t}^{(1)} \gets \*x_{t-1,w}$
            \STATE Draw  $\tilde{\*v}_{t-1}$ from the posterior $p(\*v_{t-1}\mid \*v_{t-1} < \*0)$
            \STATE $\*x_{t}^{(2)} \gets \argmax_{\*x \in \cX} \alpha(\*x)$ based on GP $f \mid \tilde{\*v}_{t-1}$
            \STATE Set $\*x_{t,w}$ and $\*x_{t,l}$ as the winner and loser of the duel between $\*x_{t}^{(1)}$ and $\*x_{t}^{(2)}$, respectively
            % \STATE Set $\*x_{t,w}$ and $\*x_{t,l}$ as the winner and loser of the duel between $\*x_{t}^{(1)}$ and $\*x_{t}^{(2)}$, respectively
            \STATE $\cD_t \gets \cD_{t-1} \cup (\*x_{t,w} \succ \*x_{t,l})$
        \ENDFOR
    \end{algorithmic}
\end{algorithm}

Algorithm~\ref{alg:HB} shows the procedure of HB.
HB iteratively selects a pair of inputs by the following two steps:
(i) Select the winner of the past duels as the first input $\*x_{t}^{(1)}$ (line 2) as with \citep{benavoli2021-preferential}.
(ii) Using the posterior distribution conditioned by the hallucination, select the input that maximizes the AF $\alpha: \cX \rightarrow \RR$ as the second point $\*x_{t}^{(2)}$ (lines 3-4).
\subsection{Choice of AF for HB}
For HB, we can use an arbitrary AF for the standard BO, e.g., EI \citep{Mockus1978-Application} and UCB \citep{Srinivas2010-Gaussian}. 
However, HB combining TS is meaningless since the sampling sequentially from $p(\*v_t | \*v_t < \*0)$ and $p(f | \*v_t)$ is just the TS.
On the other hand, we empirically observed that AFs proposed for preferential BO, such as bivariate EI \citep{nielsen2015-perception} and maximally uncertain challenge (MUC) \citep{fauvel2021-efficient}, which aims for appropriate diversity by considering the correlation between $\*x^{(1)}_t$ and $\*x^{(2)}_t$, should not be integrated into HB.
This is because, in HB, diversity is achieved by hallucination-based randomization.
Thus, HB combining bivariate EI or MUC results in over-exploration.
The empirical comparison for the HB methods combining bivariate EI, MUC, and the mean maximization are shown in Appendix~\ref{app:comparison_AF}.
Consequently, we use EI and UCB in our experiments.

Recently, several AFs for the standard BO have been proposed, e.g., predictive entropy search~\citep{Hernandez2014-Predictive}, max-value entropy search~\citep{Wang2017-Max}, and improved randomized UCB~\citep{Takeno2023-randomized}.
Confirming the performance of the HB method combined with these AFs is a possible future direction for this research.

% We employed Gibbs sampling \cite{li2015efficient} for the sampling from the truncated MVN $p(\*v_{t-1} \mid \*v_{t-1} < \*0)$, whose details are shown in Appendix~\ref{sec:Gibbs_sampling}.
% %
% Although the sampling $\tilde{\*v}_{t-1}$ needs MCMC, HB needs only one sample whose sampling is sufficiently fast (See Appendix~\ref{app:comparison_sampling} for a computational time of MCMC).
% %
% Thus, HB is highly efficient compared to \citep{benavoli2021-preferential,benavoli2021-unified}, which need many MC samples.
% %
% HB is related to the kriging believer~\cite{Shahriari2016-Taking}, which is a well-known heuristic in parallel BO literature, in the sense that some additional conditioning to the posterior is performed.
% We provide the detailed discussion in Appendix~\ref{app:diff_hallucination_kriging}.

\subsection{Relation to Existing Methods}
\label{sec:relation_TS}

HB and TS have a relationship in the perspective of using a random sample from the posterior.
HB used the hallucination from $p(\*v_t | \*v_t < \*0)$.
We can interpret that TS uses the most uncertain hallucination, i.e., sample path from $p(f | \*v_t < \*0)$.
Thus, the uncertainty of TS is very large, and TS often results in over-exploration.
On the other hand, the hallucination $\tilde{\*v}_t$ is the least uncertain random variable in the random variables, which reduces the posterior to the GP.
By this difference, HB has lower uncertainty than TS and alleviates over-exploration, particularly when the input dimension is large.

HB also has a relationship to the kriging believer \citep{Shahriari2016-Taking}, which is a well-known heuristic in the literature of parallel BO.
This relation is discussed in Appendix~\ref{app:diff_hallucination_kriging}.

%% file: manuscripts/5_related.tex
\section{Related Work}
\label{sec:related_work}

Although dueling bandit \citep{yue2012-karmed} considers online learning from the duels, it does not consider the correlation between arms in general, as reviewed in \citep{sui2018-advancements}.
In contrast, preferential BO~\citep{brochu2010-tutorial} aims for more efficient optimization using the preferential GP, which can capture a flexible correlational structure in an input domain.
% Preferential BO~\citep{brochu2010-tutorial} is a promising approach widely used for problems such as human-in-the-loop optimization.
%
Since the true posterior is computationally intractable, most prior works employed Gaussian approximation-based preferential GP models \citep{chu2005-preference,chu2005-extensions}.
Typically, preferential BO sequentially duels $\*x_t^{(1)}$, which is the most exploitative point (e.g., maxima of the posterior mean), and $\*x_t^{(2)}$, which is the maxima of some AFs.

The design of AFs for $\*x_t^{(2)}$ is roughly twofold.
First approaches \citep{brochu2010-tutorial,siivola2021-preferential} use the AFs for standard BO, which assumes that the direct observation $y$ can be obtained (e.g., EI \citep{Mockus1978-Application}).
However, this approach selects similar duels repeatedly since the preferential GP model's variance is hardly decreased due to the weakness of the information obtained from the duel.
Second approach considers the relationship between $\*x_t^{(1)}$ and $\*x_t^{(2)}$.
For example, bivariate EI \citep{nielsen2015-perception} uses the improvement from $f(\*x_t^{(1)})$, and MUC \citep{gonzalez2017preferential,fauvel2021-efficient} employed the uncertainty of the probability that $\*x_t^{(2)}$ wins $\*x_t^{(1)}$.
Considering the correlation to $\*x_t^{(1)}$, these AFs are expected to explore appropriately.
However, as discussed in Section~\ref{sec:comparison_PGP}, Gaussian approximation (LA and EP) can be inaccurate for evaluation of the joint distribution of $f(\*x_t^{(1)})$ and $f(\*x_t^{(2)})$.
%
% Therefore, these AFs using Gaussian approximation can deteriorate.

% To avoid the problem of Gaussian approximation, \citet{benavoli2021-preferential} proposed AFs, defined as the MC estimator from the true posterior, skew GP.
Instead of using Gaussian approximation, \citet{benavoli2021-preferential} directly estimates the true posterior using MCMC.
%
% Furthermore, they have shown that the sample path from skew GP can be generated using the sampling from multivariate normal (MVN) and truncated MVN (TMVN).
% Although these AFs are promising under a sufficient number of MC samples, MCMC for TMVN can require heavy computational time.
However, MCMC-based AFs require heavy computational time, which can be a bottleneck in practical applications.
% Their method need a sufficient number of MC samples from multivariate normal (MVN) and truncated MVN (TMVN) to generate sample path from skew GP, which leads to heavy computational time.
%
%
% Particularly in the application involving human interaction, the computational time, which can be user waiting time, is critical.

Another well-used AF for preferential BO is TS \citep{Russo2014-learning}.
\citet{gonzalez2017preferential} proposed to select $\*x_t^{(1)}$ by TS, and \citet{siivola2021-preferential,sui2017multi} proposed to select both $\*x_t^{(1)}$ and $\*x_t^{(2)}$ by TS, both of which are based on Gaussian approximation.
\citet{benavoli2021-preferential} also proposed to select $\*x_t^{(2)}$ by TS, which is generated from skew GP.
Although the merit of Gaussian approximation is low computational time, the posterior sampling from skew GP only once is sufficiently fast (comparable to EP).
Therefore, the advantage of Gaussian approximation for TS-based AF is limited.
In Section~\ref{sec:exp}, we will show that TS can cause over-exploration when the input dimension is large in preferential BO, as with the standard BO reviewed in \citep{Shahriari2016-Taking}.
% Thus, at first glance, TS using the skew GP seems promising in both computational and sample complexity.
%
% However, TS-based preferential BO has the problem of over-exploration, particularly when an input dimension is large, which is known in the standard BO literature \citep{Shahriari2016-Taking}, which we will show in Section~\ref{sec:exp}.

%% file: manuscripts/6_experiments.tex
\section{Comparison of Preferential BO Methods}
\label{sec:exp}

We investigate the effectiveness of the proposed preferential BO method through comprehensive numerical experiments.
We employed the 12 benchmark functions.
In this section, we show the results for the 8 functions, called the Branin, Holder table, Bukin, Eggholder, Ackley, Harmann3, Hartmann4, and Hartmann6; others are shown in Appendix~\ref{app:additional_exp}.
%
% Our experimental codes are publicly available at \url{https://github.com/CyberAgentAILab/preferentialBO}.

We performed HB combining EI \cite{Mockus1978-Application} and UCB \cite{Srinivas2010-Gaussian}, denoted as HB-EI and HB-UCB, respectively.
We employed the baseline methods, LA-EI\citep{brochu2010-tutorial}, EP-EI \cite{siivola2021-preferential}, EP-MUC~\cite{fauvel2021-efficient}, EP-TS-MUC~\citep{gonzalez2017preferential}\footnote{This method was originally proposed as ``dueling TS,'' whose name is same as \citep{benavoli2021-preferential}. To distinguish them, we denote TS-MUC since this method selects the first input by TS and the second input by MUC. Furthermore, although \citet{gonzalez2017preferential} used the GPC model, we use the EP-based preferential GP model to concentrate on the difference from the AF.}, DuelTS~\cite{benavoli2021-preferential}, DuelUCB~\cite{benavoli2021-preferential}, and EIIG\footnote{
\citet{benavoli2021-preferential,benavoli2021-unified} proposed EIIG that is EI minus information gain (IG), which is probably a typo since both EI and IG should be large. Thus, we used a modified one that EI plus IG.
}
\cite{benavoli2021-preferential}, where DuelTS, DuelUCB, and EIIG were based on skew GP and other methods with Suffix ``EP'' and ``LA'' employed EP and LA, respectively.
The first input $\*x^{(1)}_t$ for EP-EI, EP-MUC, and LA-EI is the maxima of the posterior mean over the training duels, and other methods, including HB, use the winner so far as $\*x^{(1)}_t$, following \citep{benavoli2020skew}.
For preferential GP models, we use RBF kernel with automatic relevance determination \citep{Rasmussen2005-Gaussian}, whose lengthscales are selected by marginal likelihood maximization per 10 iterations, and set fixed noise variance $\sigma^2_{\rm noise}=10^{-4}$.
We use marginal likelihood estimation by LA, which is sufficiently precise if $\sigma^2_{\rm noise}$ is small, as discussed in \citep[][Section~7.2]{Kuss2005-assessing}.
We show the accuracy of the marginal likelihood estimations of LA and EP in Appendix~\ref{app:additional_exp}.
Other detailed settings are shown in Appendix~\ref{app:exp_settings}.

Figure~\ref{fig:regret} show the results for (a) computational time and (b) iterations.
As a performance measure, we used the regret defined as $f(\*x_*) - f(\tilde{\*x}_t)$, where $\tilde{\*x}_t$ is a recommendation point at $t$-th iteration, for which we used $\*x^{(1)}_t$.
We report the mean and standard error of the regret over 10 random initialization, where the initial duel is $3d$ uniformly random input pair.

\begin{figure*}[!th]
    \centering
    \includegraphics[width=0.98\linewidth]{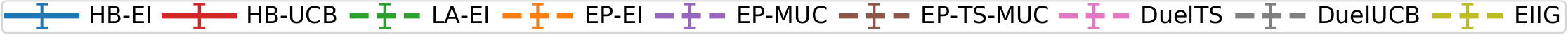}

    \centering
    \includegraphics[width=0.02\linewidth]{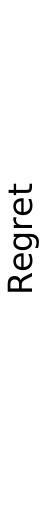}
    \includegraphics[width=0.235\linewidth]{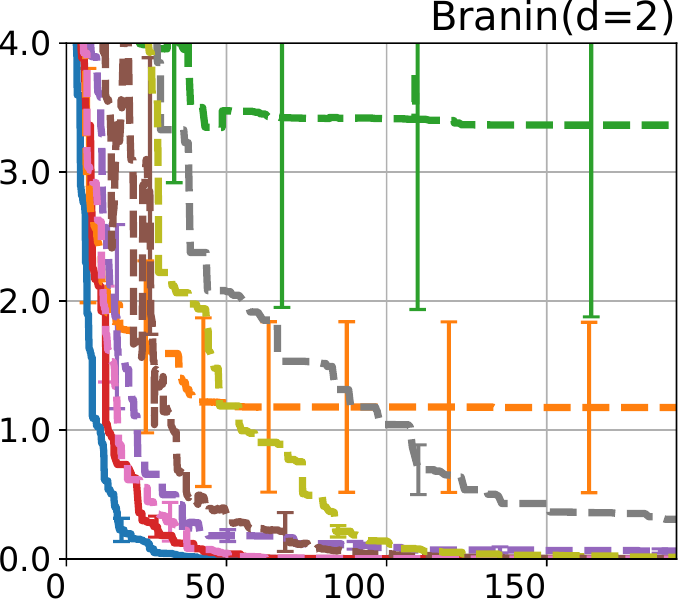}
    \includegraphics[width=0.235\linewidth]{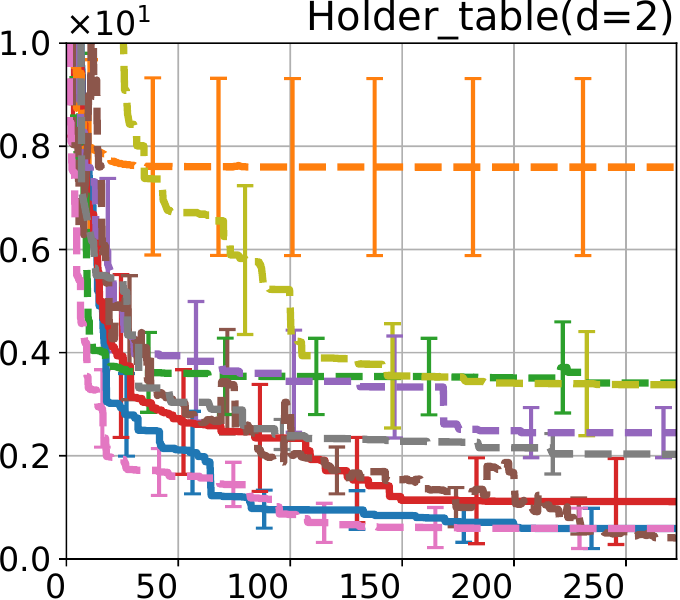}
    \includegraphics[width=0.235\linewidth]{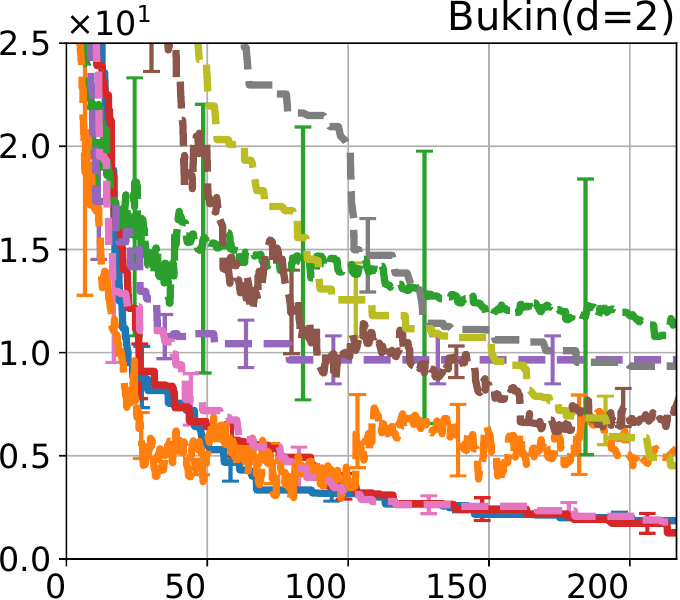}
    \includegraphics[width=0.235\linewidth]{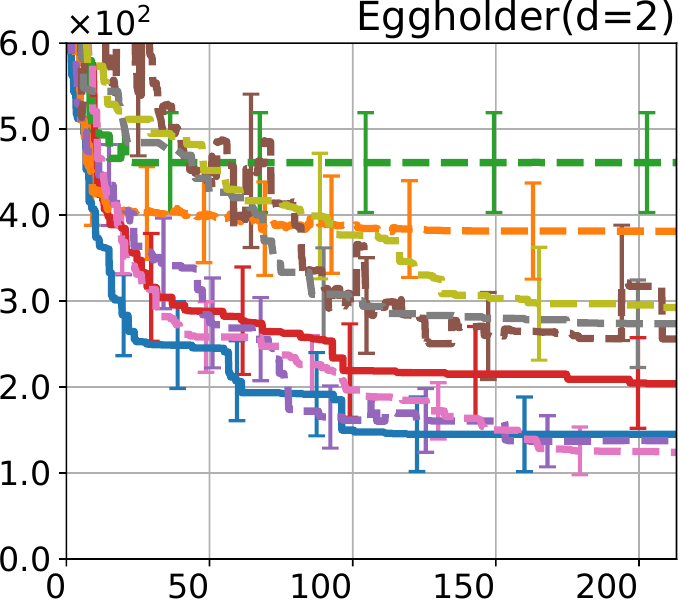}
    \hfill
    \centering
    \subfigure[Computational time (sec)]{
        \includegraphics[width=0.02\linewidth]{manuscripts/fig_new_compress/regret_axis_label-min.pdf}
        \includegraphics[width=0.235\linewidth]{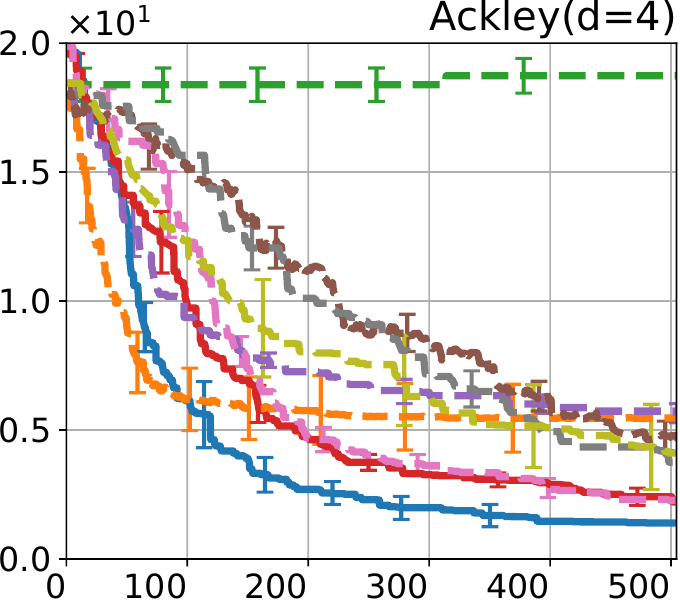}
        \includegraphics[width=0.235\linewidth]{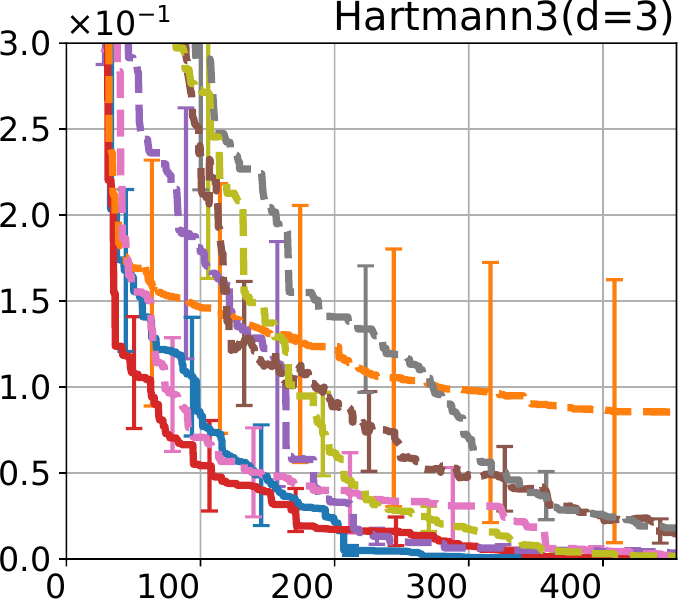}
        \includegraphics[width=0.235\linewidth]{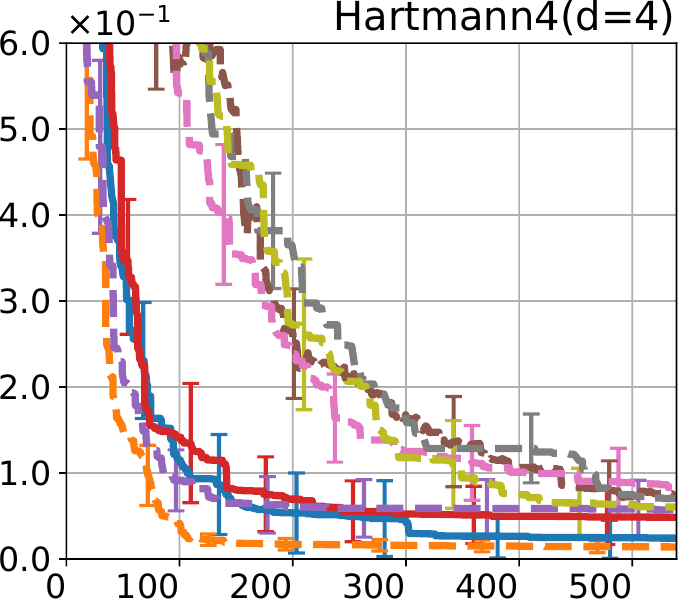}
        \includegraphics[width=0.235\linewidth]{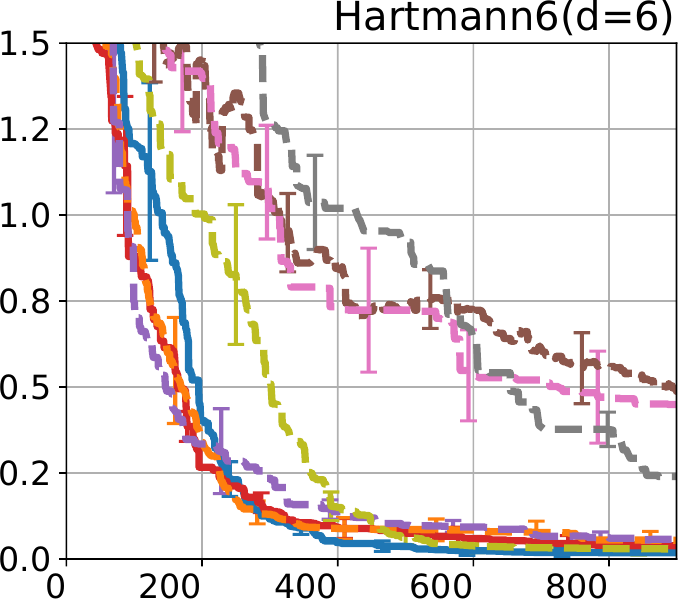}
    }
    \hfill
    \centering
    \includegraphics[width=0.02\linewidth]{manuscripts/fig_new_compress/regret_axis_label-min.pdf}
    \includegraphics[width=0.235\linewidth]{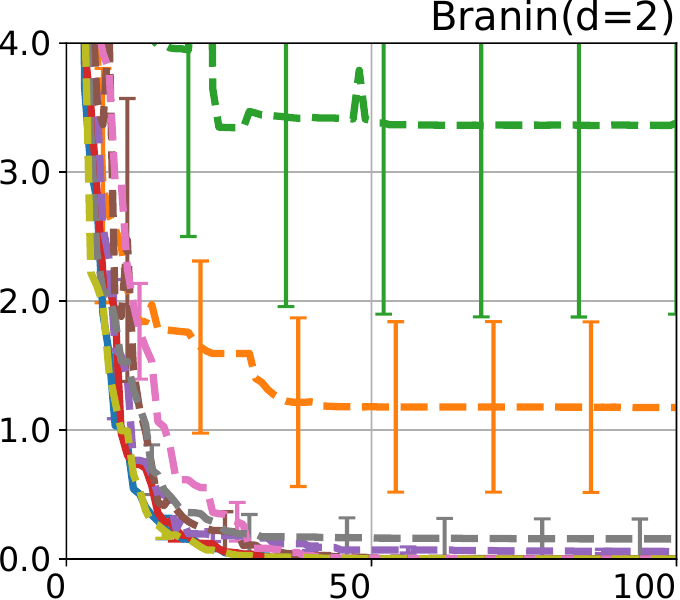}
    \includegraphics[width=0.235\linewidth]{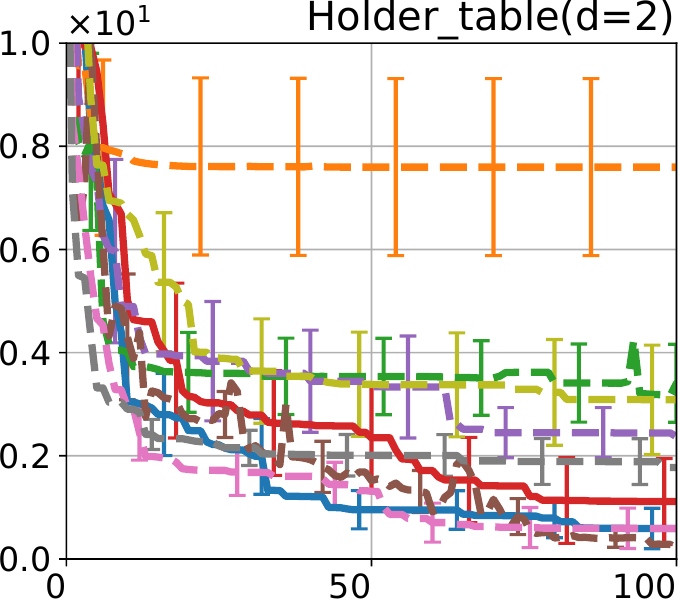}
    \includegraphics[width=0.235\linewidth]{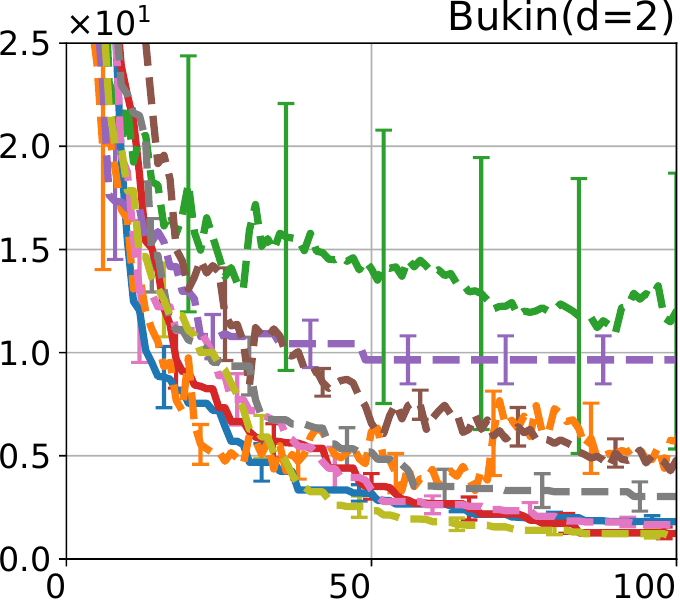}
    \includegraphics[width=0.235\linewidth]{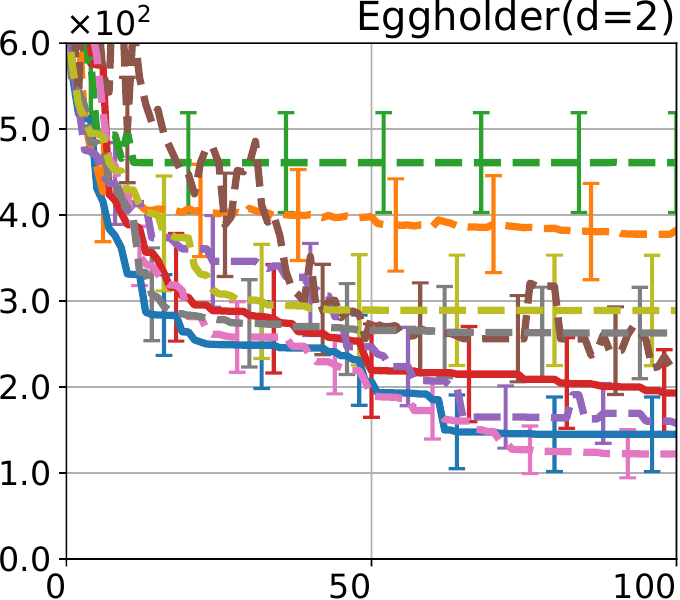}
    \hfill
    \subfigure[Iteration]{
        \includegraphics[width=0.02\linewidth]{manuscripts/fig_new_compress/regret_axis_label-min.pdf}
        \includegraphics[width=0.235\linewidth]{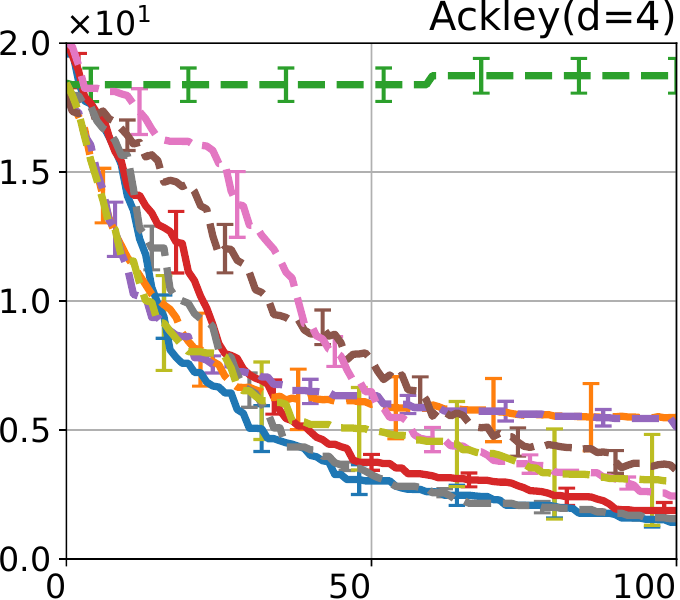}
        \includegraphics[width=0.235\linewidth]{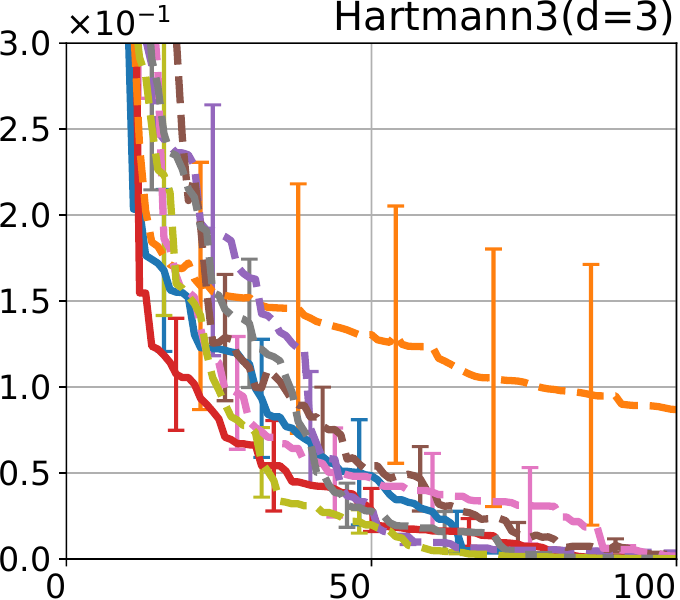}
        \includegraphics[width=0.235\linewidth]{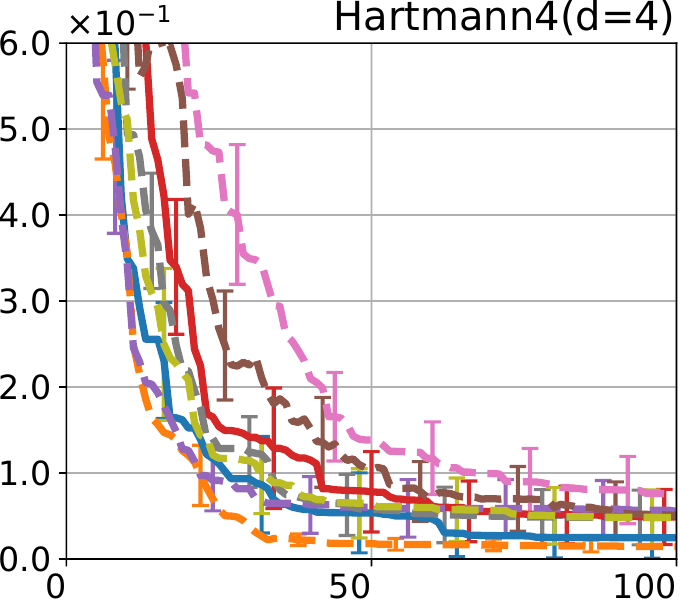}
        \includegraphics[width=0.235\linewidth]{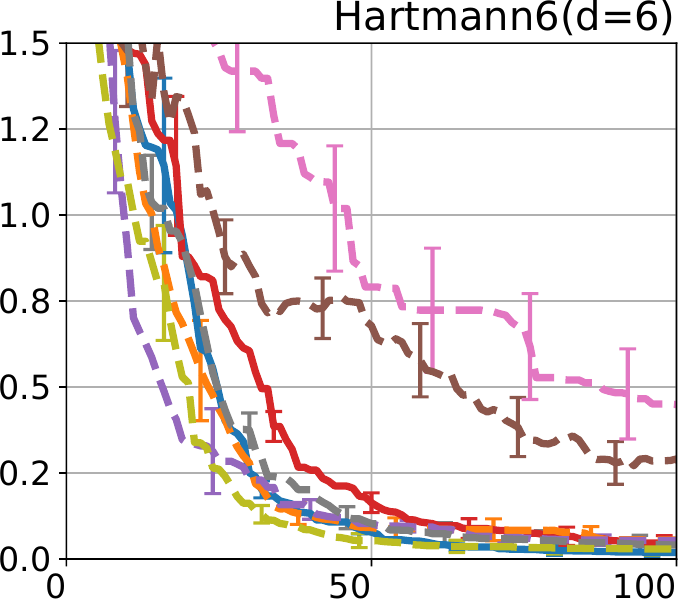}
    }
    
    \caption{
    Average and standard error of the regret of the proposed methods (HB-EI and HB-UCB) and the state-of-the-art preferential BO methods. 
    The horizontal axis represents (a) the computational time (sec) and (b) the number of iterations. 
    The vertical axis represents the regret, which is the smaller, the better it is.}
    % Suffix ``-L'' in the method name indicates using Laplace approximation.}
    \label{fig:regret}
\end{figure*}

First, we focus on the comparison against Gaussian approximation-based methods.
LA-EI is the worst method due to the poor approximation accuracy.
Note that, since we set the plotted interval for the small regret to focus the differences between HB and baseline methods except for LA-EI, LA-EI is out of the range in Hartmann 3, 4, and 6 functions.
The regret of EP-EI, which selects the duel between the maximums of the posterior mean and EI, is often stagnated due to over-exploitation.
The regrets of EP-MUC and EP-TS-MUC also stagnate in the Bukin and Ackley functions since EP is often inaccurate in predicting the joint distribution of several outputs, as shown in Section~\ref{sec:comparison_PGP}.
Furthermore, EP-TS-MUC deteriorates in the high-dimensional functions such as Hartmann 4 and 6 functions due to over-exploration from the nature of TS.
Overall, incorporating the skewness of the true posterior, HB-EI and HB-UCB outperformed Gaussian approximation-based methods in Figure~\ref{fig:regret} (b).
Furthermore, the same tendency can also be confirmed in Figure~\ref{fig:regret} (a) since the computational time of HB-EI and HB-UCB is comparable with Gaussian approximation.

Second, DuelUCB and EIIG are MCMC-based AFs based on skew GP.
Therefore, although they often show rapid convergence in Figure~\ref{fig:regret}(b) because of the use of skew GP, they show slow convergence in Figure~\ref{fig:regret}(a) due to the heavy computational time of MCMC.
Hence, HB-EI and HB-UCB are comparable to and outperformed MCMC-based AFs in Figure~\ref{fig:regret}(b) and (a), respectively, which suggest the practical effectiveness of the proposed methods.

Last, since DuelTS requires only one sample path, its performance is not largely changed in Figure~\ref{fig:regret} (a) and (b).
%
% On the other hand, TS often results in over-exploration in the standard BO literature \citep{Shahriari2016-Taking}, particularly when the input dimension is large.
%
Although the regret of DuelTS converges rapidly in low-dimensional functions, it converges considerably slowly in relatively high-dimensional functions, such as Hartmann4 and Hartmann6, in which the proposed methods clearly outperformed DuelTS.
In our experiments, we confirmed that TS results in over-exploration in relatively high-dimensional functions, which is the well-known problem of TS for the standard BO, as reviewed in \citep{Shahriari2016-Taking}.

%% file: manuscripts/7_conclusion.tex
\section{Conclusion}
Towards building a more practical preferential BO, this work has made the following important progresses.
%
% We first identified the significance of considering skewness in preferential BO by showing the poor performance of the Gaussian approximation-based methods such as Laplace approximation and expectation propagation.
%
We first revealed the poor performance of the Gaussian approximation for skew GP, such as Laplace approximation and expectation propagation, despite being the method most widely used in preferential BO.
This result motivated us to improve the MCMC-based estimation of the skew GP, for which we showed the practical effectiveness of Gibbs sampling and derived the low variance MC estimators.
%
% Finally, we developed a new preferential BO method that achieves both computational efficiency and superior optimization performance, whose effectiveness was verified through extensive numerical experiments.
Finally, we developed a new preferential BO method that achieves both high computational efficiency and low sample complexity, and then verified its effectiveness through extensive numerical experiments.
The regret analysis of the proposed method is one of the interesting future works.

%% file: manuscripts/9_appendix.tex
\section{Proof of Proposition~\ref{prop:cond_SGP}}
\label{app:proof}

We here prove Proposition~\ref{prop:cond_SGP}.
The joint posterior distribution of $\*f_{\rm tes}$ and $\*v_t$ is truncated MVN, in which $\*v_t$ is truncated above at $\*0$, as follows:
\begin{align*}
    p(\*f_{\rm tes}, \*v_t | \*v_t < \*0)
    &= \frac{\Pr(\*v_t < \*0 | \*f_{\rm tes}, \*v_t) p(\*f_{\rm tes}, \*v_t) }{\Pr(\*v_t < \*0)}.
\end{align*}
Then, we can see that $\Pr(\*v_t < \*0 | \*f_{\rm tes}, \*v_t) = \mathbbm{1} \{ \*v_t < \*0 \}$, where $\mathbbm{1} \{ \*v_t < \*0 \} = 1$ if $\*v_t < \*0$, and 0 otherwise.
Hence, we can obtain
\begin{align*}
    p(\*f_{\rm tes}, \*v_t | \*v_t < \*0)
    &= \frac{\mathbbm{1} \{ \*v_t < \*0 \} p(\*f_{\rm tes}, \*v_t) }{\Pr(\*v_t < \*0)} \\
    &= p(\*f_{\rm tes} | \*v_t) \frac{\mathbbm{1} \{ \*v_t < \*0 \} p(\*v_t) }{\Pr(\*v_t < \*0)}.
\end{align*}
Furthermore, we know that $p(\*v_t | \*v_t < \*0) = \frac{\mathbbm{1} \{ \*v_t < \*0 \} p(\*v_t) }{\Pr(\*v_t < \*0)}$.
Therefore, we derive 
\begin{align*}
    p(\*f_{\rm tes}, \*v_t | \*v_t < \*0)
    &= p(\*f_{\rm tes} | \*v_t) p(\*v_t | \*v_t < \*0),
\end{align*}
which shows $p(\*f_{\rm tes} | \*v_t) = p(\*f_{\rm tes} | \*v_t, \*v_t < \*0)$.
Since $\*f_{\rm tes}$ is arbitrary, $f \mid \*v_t$ is a GP.

We can obtain other proof.
From the property of truncated MVN \citep[][Conclusion~5]{Horrace2005-Some}, the conditional distribution of truncated MVN is truncated MVN keeping the original truncation, in which the parameters can be computed as with usual MVN.
Therefore, since remained variable $\*f_{\rm tes}$ is not truncated, the conditional distribution $p(\*f_{\rm tes} \mid \*v_t, \*v_t < \*0)$ is MVN:
\begin{align*}
    \*f_{\rm tes} \mid \*v_t, \*v_t < \*0
    \sim \cN(\*\mu_{{\rm tes} | \*v }, \*\Sigma_{{\rm tes} | \*v}),
\end{align*}
which is equivalent to the distribution $p(\*f_{\rm tes} \mid \*v_t)$.
Hence, we can see that $f \mid \*v_t$ is a GP.

%%%%%%%%%%%%%%%%%%%%%%%%%%%%%%%%%%%%%%%%%%%%%%%%%%%%%%%%%%%%%%%%%%%%%%%%%%%%%%%%%%%%%%%%%%%%%%%%%%%
\section{Detailed Derivation of Statistics for Skew GP}
\label{app:detailed_estimators}

\paragraph{Variance:}
From the law of total variance, we can decompose the variance as follows:
\begin{align*}
    \VV_{f | \*v_t < \*0}[\*f_{\rm tes}] 
    &= \VV_t[ \EE_{f | \*v_t} [\*f_{\rm tes}]] + \EE_t[\VV_{f | \*v_t}[\*f_{\rm tes}]] \\
    &= \VV_t[ \*\mu_{{\rm tes} | \*v }] + \EE_t[\*\Sigma_{{\rm tes} | \*v}] && (\because \text{Proposition~\ref{prop:cond_SGP}}) \\
    &= \VV_t[ \*\mu_{{\rm tes} | \*v }] + \*\Sigma_{{\rm tes} | \*v}. && (\because \text{$\*\Sigma_{{\rm tes} | \*v}$ does not depend on $\*v_t$.})
\end{align*}

\paragraph{CDF of $f(\*x)$:}
We can derive as follows:
\begin{align*}
    \myPr \bigl( [\*f_{\rm tes}]_i \leq c  | \*v_t < \*0 \bigr) 
    &= \EE_t \left[ \Pr \bigl( [\*f_{\rm tes}]_i \leq c | \*v_t, \*v_t < \*0 \bigr) \right] \\
    &= \EE_t \left[ \Phi \left( \frac{c - [\*\mu_{{\rm tes} | \*v}]_i}{ \sqrt{[\*\Sigma_{{\rm tes} | \*v}]_{i,i} }} \right) \right].&& (\because \text{Proposition~\ref{prop:cond_SGP}})
\end{align*}

\paragraph{Probability of Duel:}
Given $\*v_t$, $[\*f_{\rm tes}]_i$ and $[\*f_{\rm tes}]_j$ follow MVN jointly.
Therefore, $[\*f_{\rm tes}]_i - [\*f_{\rm tes}]_j | \*v_t$ also follows the Gaussian distribution $\cN([\*\mu_{{\rm tes} | \*v}]_i - [\*\mu_{{\rm tes} | \*v}]_j, [\*\Sigma_{{\rm tes} | \*v}]_{i,i} + [\*\Sigma_{{\rm tes} | \*v}]_{j,j} - 2 [\*\Sigma_{{\rm tes} | \*v}]_{i,j})$.
Then, we can obtain the estimator as follows:
\begin{align*}
    \myPr \bigl( [\*f_{\rm tes}]_i  \leq [\*f_{\rm tes}]_j | \*v_t < \*0 \bigr) 
    &= \myPr \bigl( [\*f_{\rm tes}]_i - [\*f_{\rm tes}]_j \leq 0 | \*v_t < \*0 \bigr) \\
    &= \EE_t \left[ \myPr \bigl( [\*f_{\rm tes}]_i - [\*f_{\rm tes}]_j \leq 0 | \*v_t, \*v_t \leq \*0 \bigr) \right] \\
    &= \EE_t \left[ \Phi \left( 
    \frac{[\*\mu_{{\rm tes} | \*v}]_j - [\*\mu_{{\rm tes} | \*v}]_i}{ \sqrt{[\*\Sigma_{{\rm tes} | \*v}]_{i,i} + [\*\Sigma_{{\rm tes} | \*v}]_{j,j} - 2 [\*\Sigma_{{\rm tes} | \*v}]_{i,j} }}
    \right) \right]. && (\because \text{Proposition~\ref{prop:cond_SGP}})
\end{align*}

\paragraph{Quantile:}
As mentioned in the main paper, the quantile estimators can be defined as an inverse function of the MC estimator of Eq.~\eqref{eq:CDF_estimator}.
Let the estimator of Eq.~\eqref{eq:CDF_estimator} be $\hat{F}_i: \RR \rightarrow [0, 1]$.
Unfortunately, since $\hat{F}_i$ is defined as the mean of the CDF of Gaussian distribution, the analytical form of $\hat{F}_i^{(-1)}$ is unknown.
Therefore, we apply the binary search to $\hat{F}_i$.
We define the initial search interval from the $M$ MC samples $\tilde{\*v}_{t, k}$, where $k \in \{1, \dots, M\}$.
From the construction of $\hat{F}_i$, we see that
\begin{align*}
    \hat{F}_i^{(-1)} (\alpha) 
    \in [\ell, u] 
    \coloneqq 
        \left[ 
            \min_k [\*\mu_{{\rm tes} | \tilde{\*v}_{t,k}}]_i + c \sqrt{[\*\Sigma_{{\rm tes} | \tilde{\*v}_{t,k}}]_{i,i} }, 
            \max_k [\*\mu_{{\rm tes} | \tilde{\*v}_{t,k}}]_i + c \sqrt{[\*\Sigma_{{\rm tes} | \tilde{\*v}_{t,k}}]_{i,i} } 
        \right],
\end{align*} 
where $\alpha \in (0, 1)$, $c = \Phi^{-1}(\alpha)$, and $\ell$ and $u$ are the most extreme value of the quantiles of $[\*f_{\rm tes}]_i | \tilde{\*v}_{t, k}$ over $k \in \{1, \dots, M\}$.
Therefore, we use $[\ell, u]$ as the initial interval.
Algorithm~\ref{alg:quantile} shows the pseudo-code.

\begin{algorithm}[t]
    \caption{Binary search for quantiles of skew GP} \label{alg:quantile}
    \begin{algorithmic}[1]
        \REQUIRE $\hat{F}_i: \RR \rightarrow [0, 1]$, the desired percentage $\alpha \in (0, 1)$, tolerance $\epsilon$
        \STATE $c \gets \Phi^{-1}(\alpha)$
        \STATE $[\ell, u] \gets \left[ \min_k [\*\mu_{{\rm tes} | \tilde{\*v}_{t,k}}]_i + c \sqrt{[\*\Sigma_{{\rm tes} | \tilde{\*v}_{t,k}}]_{i,i} }, \max_k [\*\mu_{{\rm tes} | \tilde{\*v}_{t,k}}]_i + c \sqrt{[\*\Sigma_{{\rm tes} | \tilde{\*v}_{t,k}}]_{i,i} } \right]$
        \WHILE{$|\hat{F}_i(\gamma) - \alpha| > \epsilon$}
            \STATE $\gamma \gets \hat{F}_i\bigl( (\ell + u) / 2 \bigr)$
            \IF{$\gamma > \alpha$}
                \STATE $u \gets (\ell + u) / 2$
            \ELSE
                \STATE $\ell \gets (\ell + u) / 2$
            \ENDIF
        \ENDWHILE
        \ENSURE $\gamma$
    \end{algorithmic}
\end{algorithm}

%%%%%%%%%%%%%%%%%%%%%%%%%%%%%%%%%%%%%%%%%%%%%%%%%%%%%%%%%%%%%%%%%%%%%%%%%%%%%%%%%%%%%%%%%%%%%%%%%%%
\section{Proof of Proposition~\ref{prop:low_variance}}
\label{app:proof_variance}

We show the general fact for the MC estimator.
Let $X$ and $Y$ be random variables, where $Y$ depends on $X$.
Furthermore, we assume the conditional expectation $\mu_{Y|X} \coloneqq \EE_{Y | X} [Y]$ can be analytically calculated from a realization of $X$.
Then, $Y$ itself and $\mu_{Y|X}$ are both unbiased estimators of $\EE[Y]$, i.e., $\EE_Y[Y] = \EE_X[\mu_{Y|X}]$.
Intuitively, since $\mu_{Y|X}$ is computed analytically with respect to the expectation of $Y | X$, it should be a better estimator than $Y$.
This intuition is justified by the law of total variance shown below:
\begin{align*}
    \VV[Y] = \VV_X[\mu_{Y|X}] + \EE_X[\VV_{Y|X}[Y]].
\end{align*}
Since $\EE_X[\VV_{Y|X}[Y]] \geq 0$, we see $\VV_X[\mu_{Y|X}] \leq \VV[Y]$.
In the case of our mean estimator, $X$, $Y$, and $\mu_{Y|X}$ correspond to $\*v_t$, $f(\*x_{i, {\rm tes}})$, and $\mu_{{\rm tes} | \*v_t}$.
This concludes the proof.

%%%%%%%%%%%%%%%%%%%%%%%%%%%%%%%%%%%%%%%%%%%%%%%%%%%%%%%%%%%%%%%%%%%%%%%%%%%%%%%%%%%%%%%%%%%%%%%%%%%
\section{Detailed Procedure of Gibbs Sampling for TMVN}
\label{app:detail_gibbs_sampling}

% Gibbs sampling is often used for the sampling of TMVN \citep{adler2008-efficient,li2015efficient,geweke1991efficient,Breslaw1994-random}.
%
Let us consider the sampling from TMVN $p(\*v | \*v < \*0)$, where original $\*v \in \RR^{n}$ follows MVN below:
\begin{align*}
    \*v \sim \cN(0, \*\Sigma),
\end{align*}
where $\*\Sigma \in \RR^{n\times n}$ is an arbitrary covariance matrix.
Let $v_j$ and $\*v_{-j}$ be $j$-th element of $\*v$ and the vector consisting of the elements except for $v_j$, respectively.
In Gibbs sampling, we repeat the sampling $v_j \mid \*v_{-j}, v_j < 0$, which follows a univariate truncated normal distribution.
The conditional distribution $v_j \mid \*v_{-j} \sim \cN(\mu_j, \sigma^{2}_j)$, where $\mu_j$ and $\sigma^2_j$ are computed efficiently by computing $\*\Sigma^{-1}$ once \citep[][Section~5.4.2]{Rasmussen2005-Gaussian}.
Algorithm~\ref{alg:GS} shows the procedure of Gibbs sampling.
% , in which $[\cdot]_j$ and $[\cdot]_{jj}$ imply the $j$-th element of the vector and $(j,j)$-th element of the matrix, respectively.
%
For the sampling from the univariate truncated normal distribution, we employed the efficient rejection sampling \citep[][Section~2.1]{li2015efficient}, which uses several proposal distributions depending on the truncation.
Note that we did not employ the transformation of $\*v$, which is proposed in \citep[][Section 2.2]{li2015efficient} to improve the mixing time.
This is because we cannot confirm the effectiveness of this transformation due to additional computational time for transformation.

% \begin{algorithm}[!th]
%     \caption{Gibbs sampling for TMVN}\label{alg:GS}
%     \begin{algorithmic}
%         \Require $\*v^{(0)}$, $\*\Sigma$
%         \State Compute $\*\Sigma^{-1}$
%         \For{$i = 1, \dots$}
%             \State $\*v^{(i)} \gets \*v^{(i-1)}$
%             \For{$j = 1, \dots, n$}
%                 \State $\mu^{(i)}_j \gets [\Sigma^{-1} \*v^{(i)}]_j / [\Sigma^{-1}]_{jj}$
%                 % \State $\sigma^{2(i)}_j \gets 1 / [\Sigma^{-1}]_{jj}$
%                 \State Set $v^{(i)}_j$ by the sampling from $\cN(\mu^{(i)}_j, \sigma^{2}_j = 1 / [\Sigma^{-1}]_{jj})$ with truncation above at $\*0$
%             \EndFor
%         \EndFor
%     \end{algorithmic}
% \end{algorithm}

\begin{algorithm}[th]
    \caption{Gibbs sampling for TMVN}\label{alg:GS}
    \begin{algorithmic}
        \REQUIRE $\*v^{(0)}$, $\*\Sigma$
        \STATE Compute $\*\Sigma^{-1}$
        \FOR{$i = 1, \dots$}
            \STATE $\*v^{(i)} \gets \*v^{(i-1)}$
            \FOR{$j = 1, \dots, n$}
                \STATE $\mu^{(i)}_j \gets [\Sigma^{-1} \*v^{(i)}]_j / [\Sigma^{-1}]_{jj}$
                % \State $\sigma^{2(i)}_j \gets 1 / [\Sigma^{-1}]_{jj}$
                \STATE Set $v^{(i)}_j$ by the sampling from $\cN(\mu^{(i)}_j, \sigma^{2}_j = 1 / [\Sigma^{-1}]_{jj})$ with truncation above at $\*0$
            \ENDFOR
        \ENDFOR
    \end{algorithmic}
\end{algorithm}

%%%%%%%%%%%%%%%%%%%%%%%%%%%%%%%%%%%%%%%%%%%%%%%%%%%%%%%%%%%%%%%%%%%%%%%%%%%%%%%%%%%%%%%%%%%%%%%%%%%
\section{Difference between Hallucination and Kriging Believer}
\label{app:diff_hallucination_kriging}

Kriging believer (KB) \citep{Shahriari2016-Taking} is a well-known heuristic for parallel BO.
KB conditions on the ongoing function evaluation by the posterior mean, and then the next batch point is selected using this GP conditioned by the posterior mean.
Some variants, called constant liar, use some predefined constant instead of the posterior mean.
Furthermore, some studies averaged the resulting AF value by the sample of the posterior normal distribution \citep[e.g., ][]{Snoek2012-Practical}.
These studies aim to guarantee the diversity of the batch points via penalization by conditioning.

On the other hand, one of the important aims of HB is to reduce the skew GP to the standard GP.
For this purpose, we conditioned the latent truncated variable $\*v_t$, which is not related to parallel BO.
Furthermore, in preferential BO, if we conditioned the constant including the posterior mean, preferential BO methods cannot consider the skewness.
Thus, the conditioning by the constant results in poor performance as with Gaussian approximation.
On the other hand, although averaging by the samples from the posterior is promising, it requires huge computational time for MCMC with respect to skew GP.
Hence, we employed the conditioning by the hallucination of $\*v_t$.
%
% The conditioning of $y_{\*x^{(1)}_t}$ is used for the same purpose of KB, i.e., the penalization.

%%%%%%%%%%%%%%%%%%%%%%%%%%%%%%%%%%%%%%%%%%%%%%%%%%%%%%%%%%%%%%%%%%%%%%%%%%%%%%%%%%%%%%%%%%%%%%%%%%%
\section{Detailed Experimental Settings}
\label{app:exp_settings}

All the details of benchmark functions are shown in \url{https://www.sfu.ca/~ssurjano/optimization.html}.

\paragraph{Comparison of Preferential GPs:} 
We fitted LA-, EP-, and MCMC-based preferential GP with RBF kernel to 50 uniformly random duels in 10 random trials.
We set $\sigma^2_{\rm noise} = 10^{-4}$, and the hyperparameters of RBF kernel are determined by marginal likelihood maximization \citep{Rasmussen2005-Gaussian}, for which we use LA-based marginal likelihood estimation.
We evaluate the estimators of the mean, mode, and the duel probabilities $\myPr_t\bigl( f(\*x) \succ f(\*x^\prime) \bigr)$ at 100 uniformly random inputs and inputs included in the training duels.
The MCMC uses 10000 MC samples, 1000 burn-in, and 10 thinning.

\paragraph{Comparison of Preferential BO Methods:}
%
% In each function, we used the RBF kernel with automatic relevance determination \cite{Rasmussen2005-Gaussian}, whose hyper-parameters are chosen using the marginal likelihood maximization using LA, which is sufficiently precise if $\sigma^2_{\rm noise}$ is small, as discussed in \citep[][Section~7.2]{Kuss2005-assessing}.
%
Although \cite{benavoli2021-preferential} originally employed LinESS for the sampling from truncated MVN, we employed Gibbs sampling in DuelTS, DuelUCB, and EIIG, as with our proposed methods, for a fair comparison.
For the parameters for Gibbs sampling, burn-in is 1000, and the MC sample size for DuelUCB and EIIG is 1000 (thinning is not performed).
Other settings for existing methods, such as the percentage for UCB, are set as with the suggestion from the original paper.
For HB-UCB, we use $\beta^{1/2} = 2$.

% For the model selection of preferential GP models, we use marginal likelihood estimation by LA, which is sufficiently precise if $\sigma^2_{\rm noise}$ is small, as discussed in \citep[][Section~7.2]{Kuss2005-assessing}.
% %
% We show the accuracy of the marginal likelihood estimations of LA and EP in Appendix~\ref{app:additional_exp}.

%%%%%%%%%%%%%%%%%%%%%%%%%%%%%%%%%%%%%%%%%%%%%%%%%%%%%%%%%%%%%%%%%%%%%%%%%%%%%%%%%%%%%%%%%%%%%%%%%%%
\section{Additional Experiments}
\label{app:additional_exp}
In this section, we provide additional experimental results.

%%%%%%%%%%%%%%%%%%%%%%%%%%%%%%%%%%%%%%%%%%%%%%%%%%%%%%%%%%%%%%%%%%%%%%%%%%%%%%%%%%%%%%%%%%%%%%%%%%%
\subsection{Marginal Likelihood Approximation}

\begin{figure}[t]
    \centering
    \includegraphics[width=\linewidth]{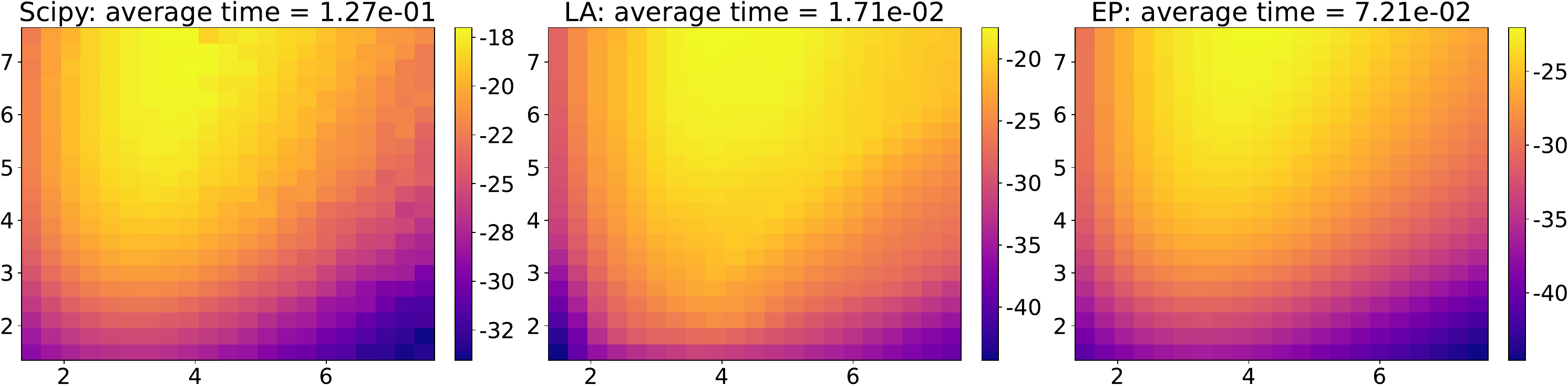}
    \caption{Log marginal likelihood estimations for Branin function by Scipy CDF of MVN, LA, and EP. Vertical and horizontal axes represent lengthscales of RBF kernel. In titles, averages of computational times are shown.}
    \label{fig:MLE}
\end{figure}

The marginal likelihood of the preferential GP model is $\Pr(\*v_t < \*0)$, the CDF of MVN.
LA and EP can provide computationally efficient approximations of this.
In the literature of GPC, \citet{Kuss2005-assessing} showed the broad evaluations for the accuracy of these approximations, which suggest that EP is accurate.
Furthermore, \citet[][Section~7.2]{Kuss2005-assessing} described that LA is also accurate if the noise variance $\sigma^2_{\rm noise} < 1$.
We here show the examples using the Branin function in Figure~\ref{fig:MLE}, in which the marginal likelihood approximations by Scipy CDF of MVN, LA, and EP are shown.
We can see that LA and EP show reasonable approximations for the marginal likelihood.
In addition, the computational time shown in the title implies the efficiency of LA.
Therefore, we employed the LA-based marginal likelihood approximation throughout the paper.

%%%%%%%%%%%%%%%%%%%%%%%%%%%%%%%%%%%%%%%%%%%%%%%%%%%%%%%%%%%%%%%%%%%%%%%%%%%%%%%%%%%%%%%%%%%%%%%%%%%
\subsection{Approximation of Preferential GP for Other Benchmark Functions}
\label{app:comparison_sampling}

We here show the additional experiments regarding preferential GP, which include the comparisons between Gibbs sampling and LinESS,  MCMC and LA and EP, and our MC estimator and the full MC estimator, using the Holder table and Hartmann6 functions.
The experimental settings are the same as the experiments for the Ackley function.

\begin{figure}[t]
    \begin{minipage}{0.49\linewidth}
        \centering
        \subfigure[Trace and autocorrelation plots of proposed MC estimator and full MC estimator \citep{benavoli2021-preferential}.]{\includegraphics[width=\linewidth]{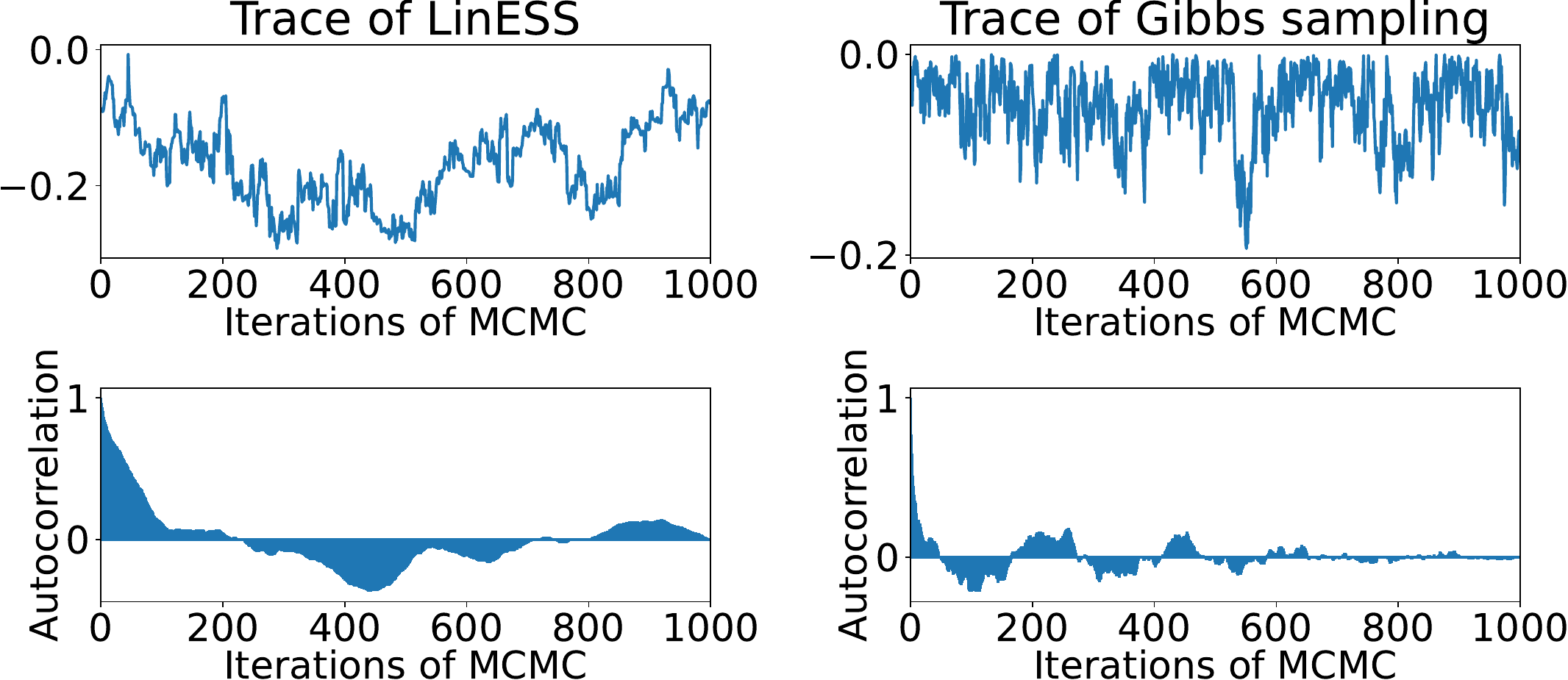}}

        \subfigure[RMSE of proposed MC estimator and full MC estimator \citep{benavoli2021-preferential} against ground truth.]{
            \centering
            \includegraphics[width=0.345\linewidth]{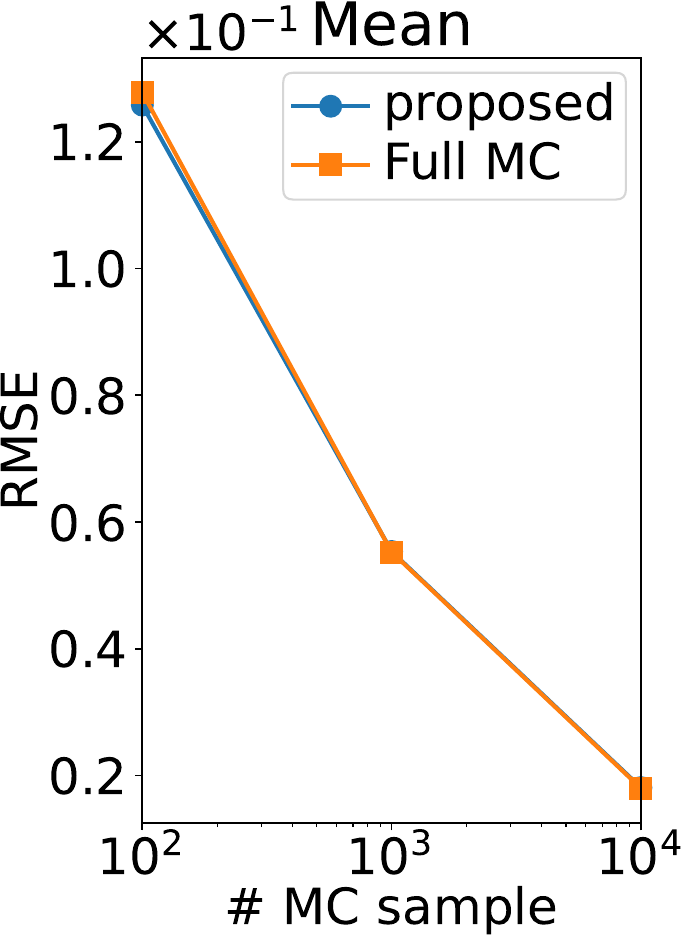}
            \includegraphics[width=0.317\linewidth]{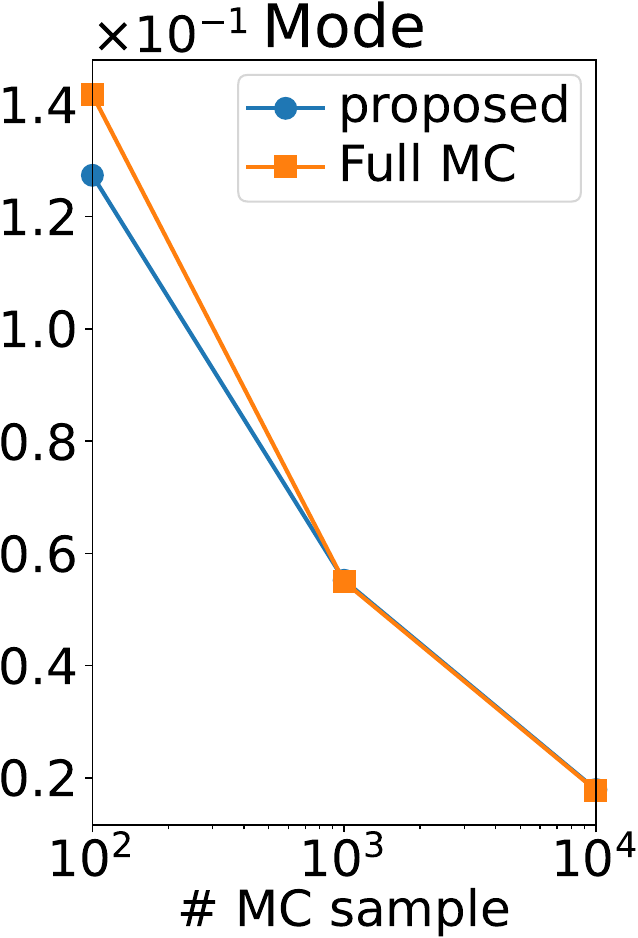}
            \includegraphics[width=0.318\linewidth]{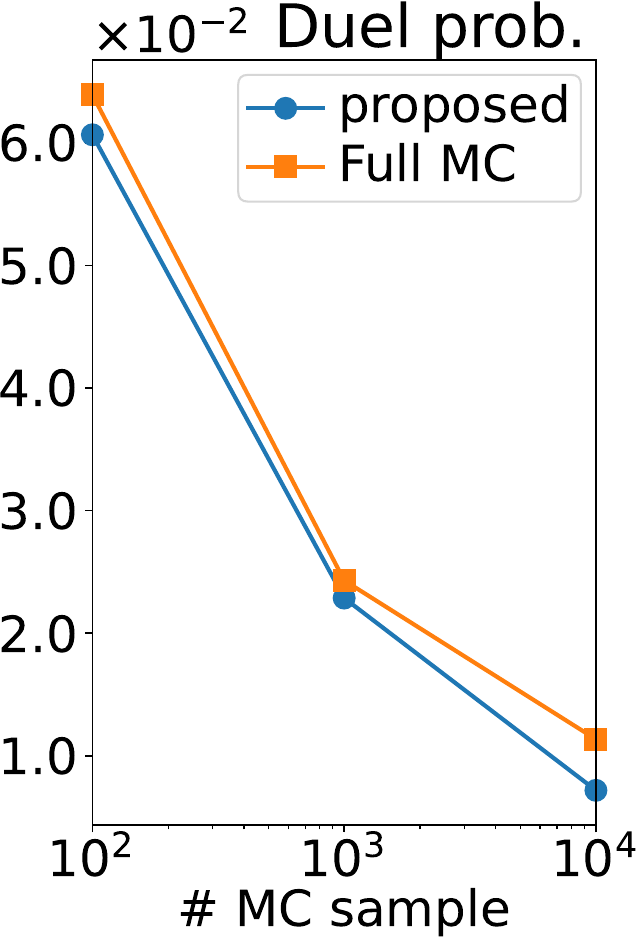}
        }
    
        \centering
        \includegraphics[width=0.32\linewidth]{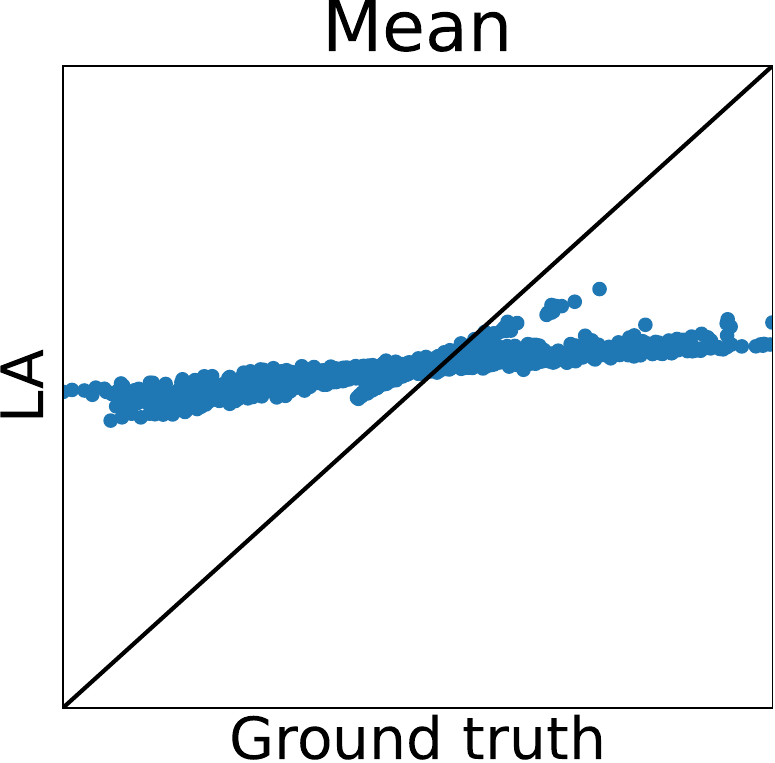}
        \includegraphics[width=0.32\linewidth]{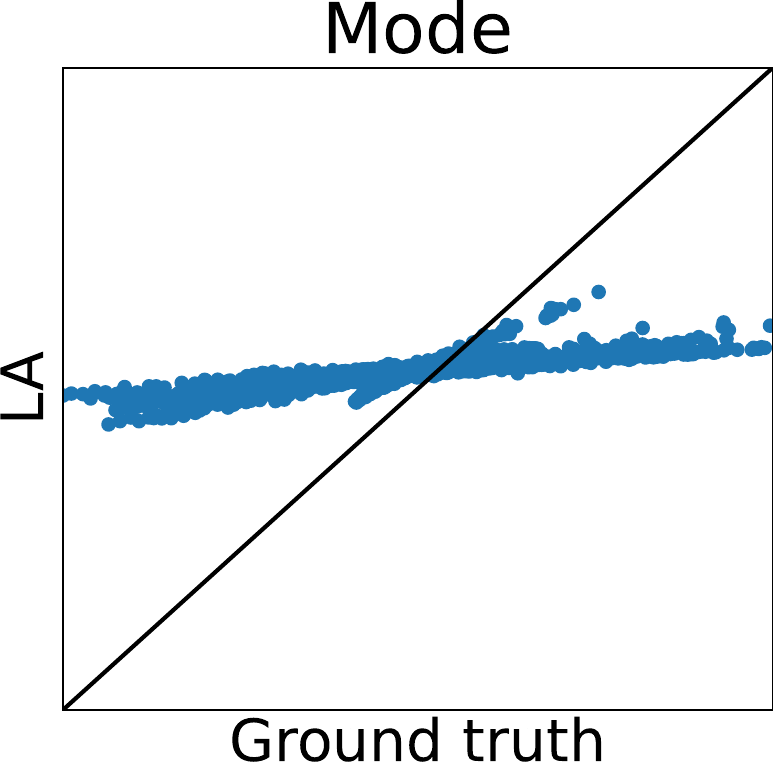}
        \includegraphics[width=0.32\linewidth]{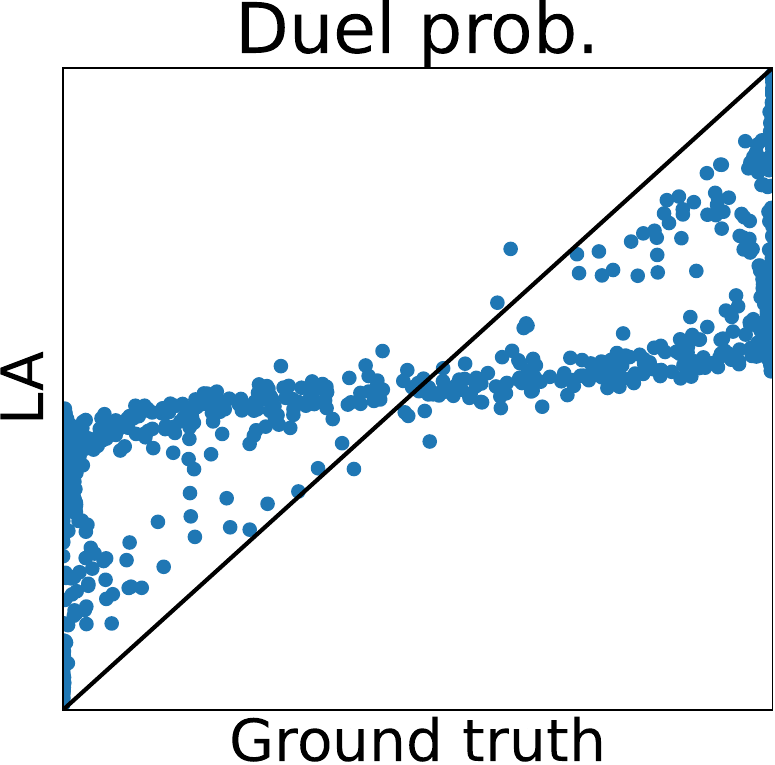}
        \subfigure[Truth vs. prediction plots for LA and EP.]{
            \centering
            \includegraphics[width=0.32\linewidth]{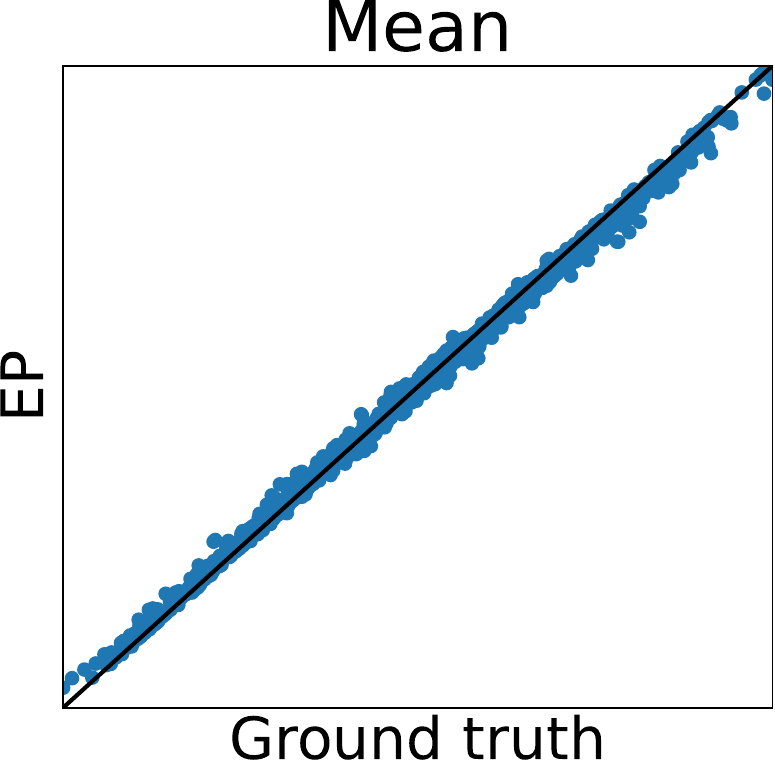}
            \includegraphics[width=0.32\linewidth]{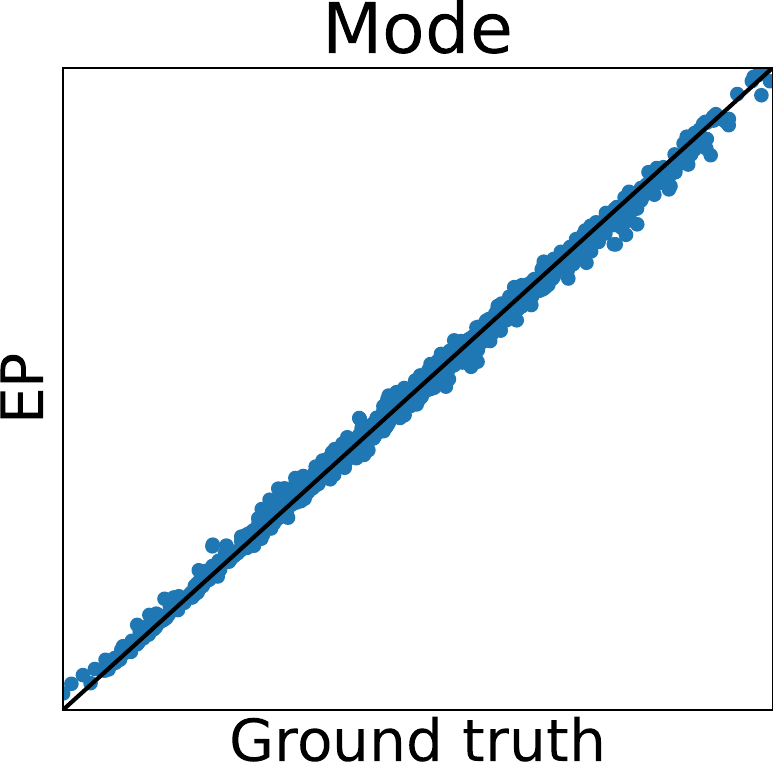}
            \includegraphics[width=0.32\linewidth]{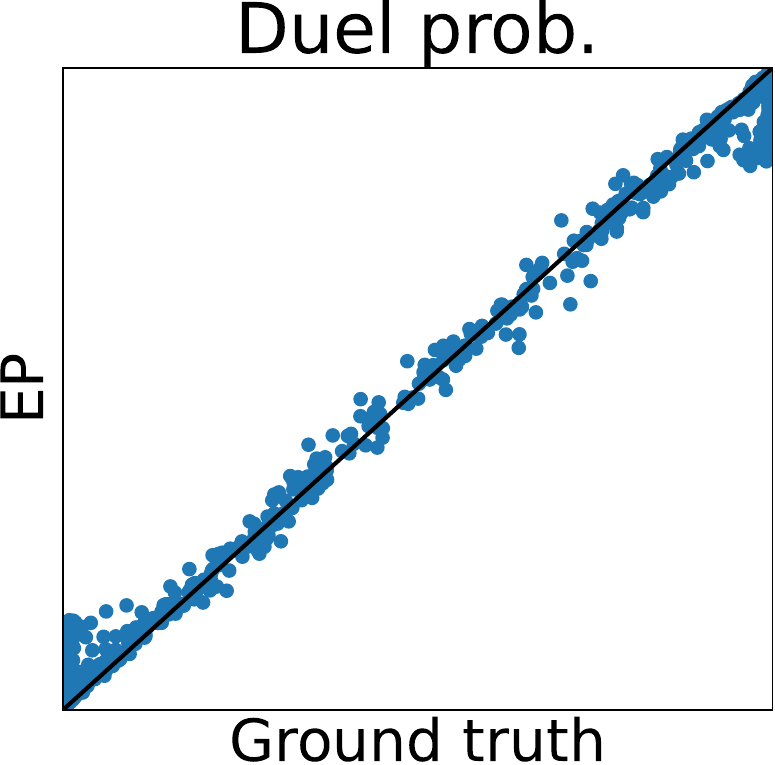}
        }
        \caption{Reuslts of the Holder table function ($d=2$).}
        \label{fig:pgp_holder_table}
    \end{minipage}
    \hspace{0.02\linewidth}
    \begin{minipage}{0.49\linewidth}
        \centering
        \subfigure[Trace and autocorrelation plots of proposed MC estimator and full MC estimator \citep{benavoli2021-preferential}.]{\includegraphics[width=\linewidth]{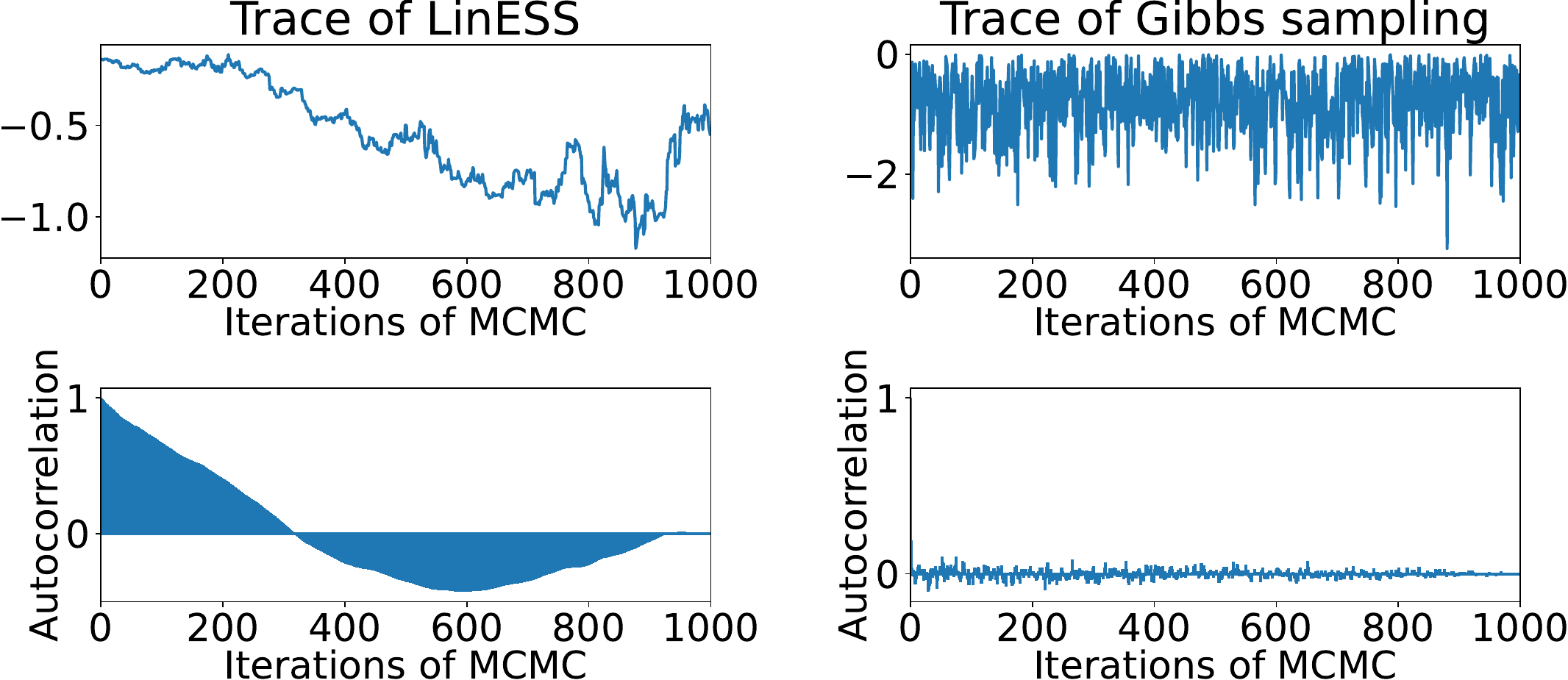}}

        \subfigure[RMSE of proposed MC estimator and full MC estimator \citep{benavoli2021-preferential} against ground truth.]{
            \centering
            \includegraphics[width=0.345\linewidth]{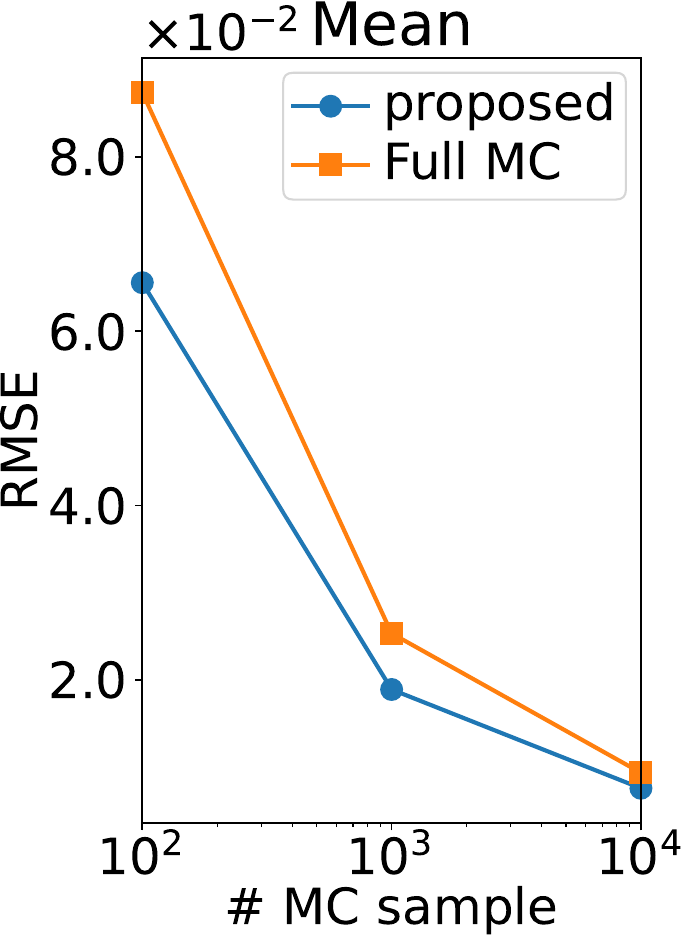}
            \includegraphics[width=0.317\linewidth]{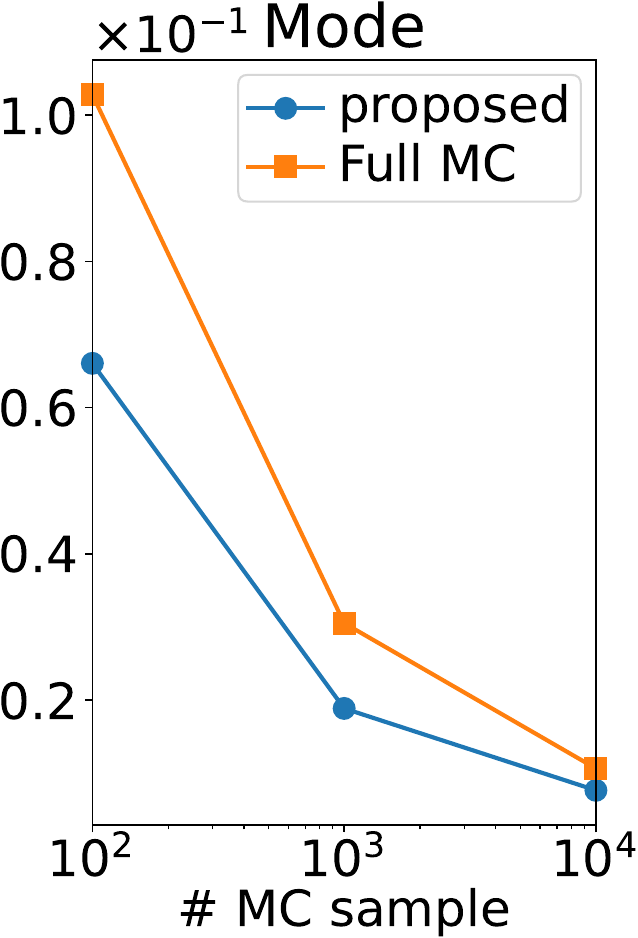}
            \includegraphics[width=0.318\linewidth]{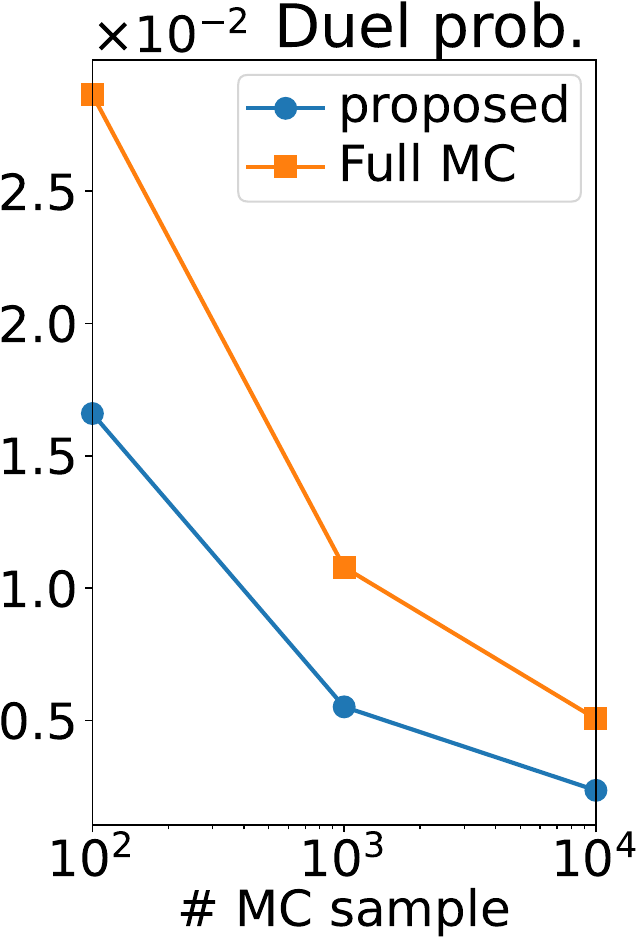}
        }
    
        \centering
        \includegraphics[width=0.32\linewidth]{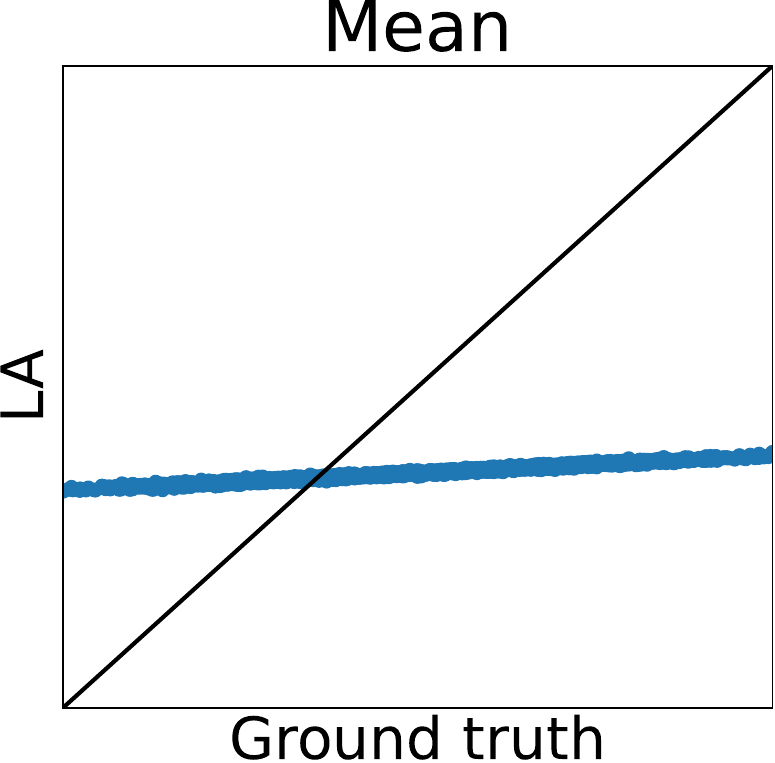}
        \includegraphics[width=0.32\linewidth]{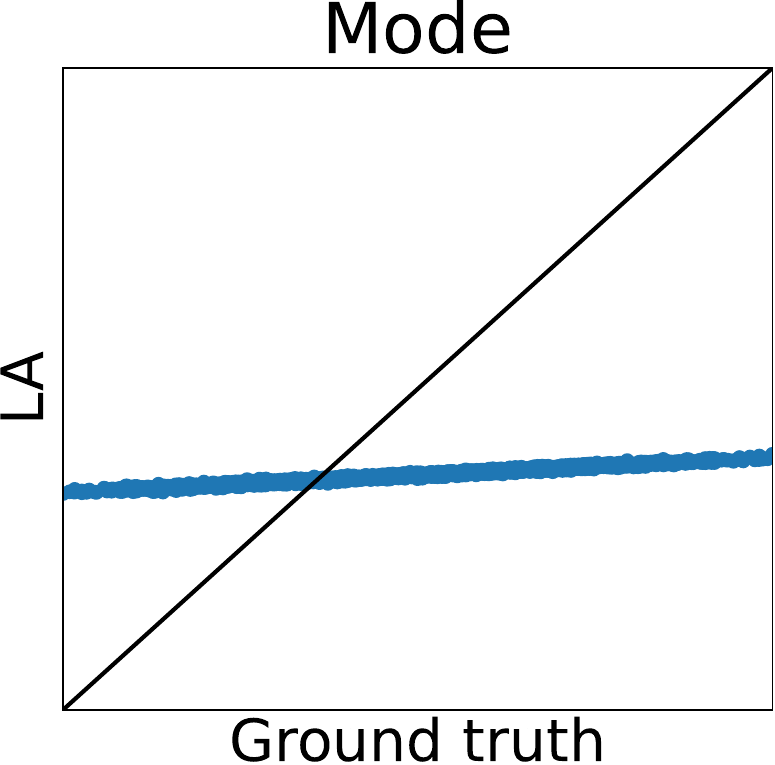}
        \includegraphics[width=0.32\linewidth]{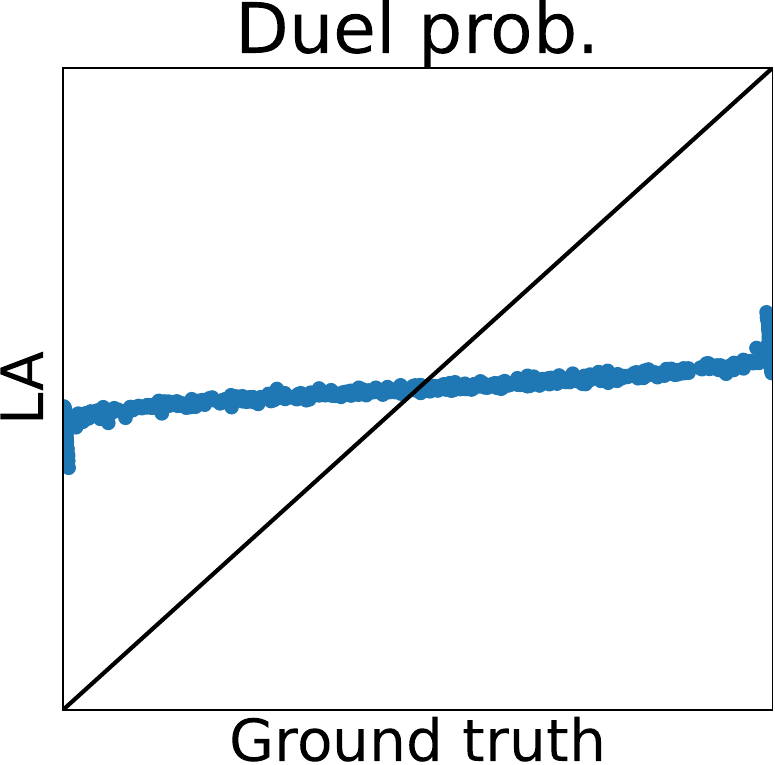}
        \subfigure[Truth vs. prediction plots for LA and EP.]{
            \centering
            \includegraphics[width=0.32\linewidth]{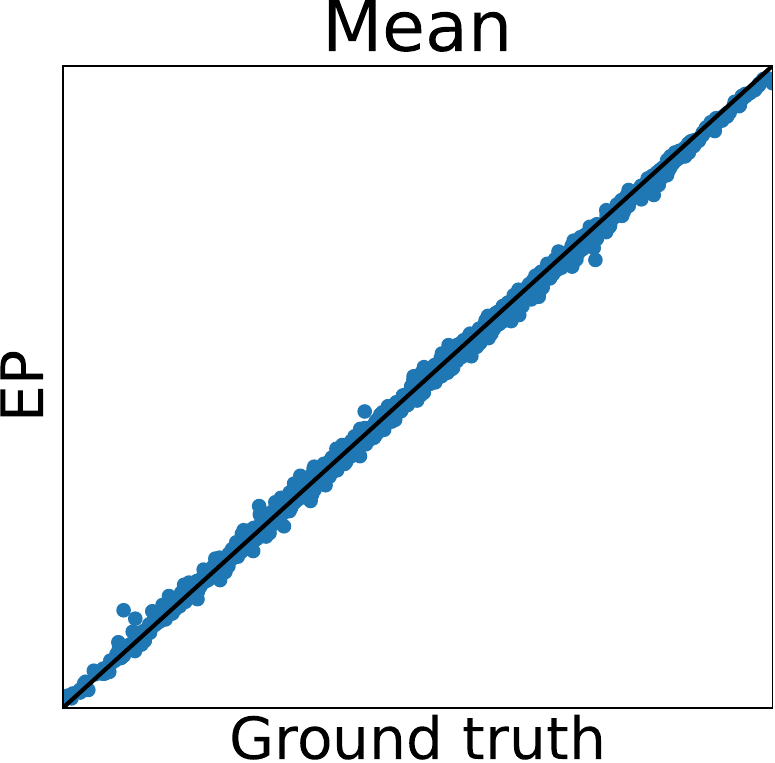}
            \includegraphics[width=0.32\linewidth]{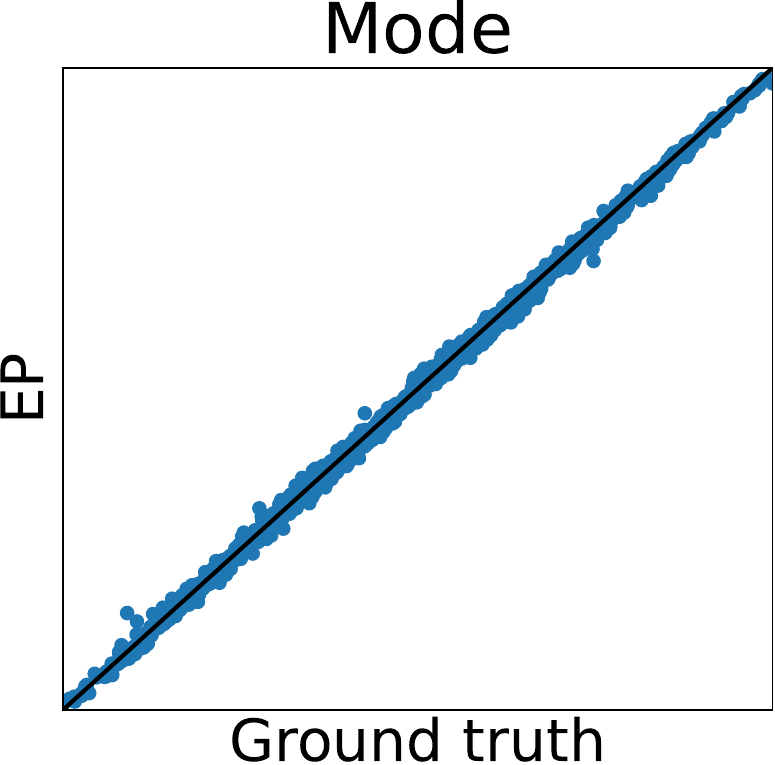}
            \includegraphics[width=0.32\linewidth]{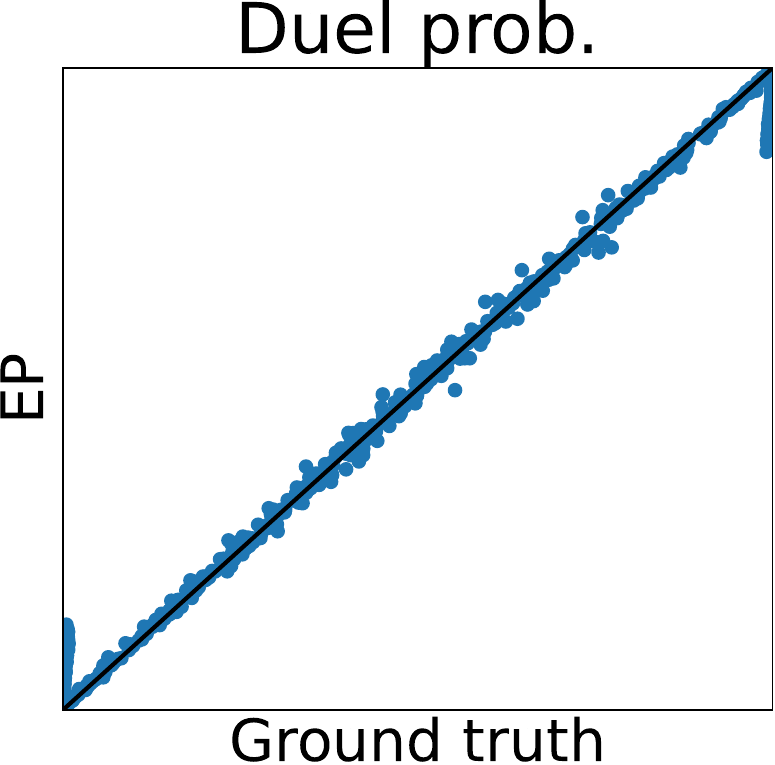}
        }
        \caption{Reuslts of the Hartmann6 function ($d=2$).}
        \label{fig:pgp_hartmann6}
    \end{minipage}
\end{figure}

Figures~\ref{fig:pgp_holder_table} and \ref{fig:pgp_hartmann6} show the results of the Holder table and Hartmann6 functions, respectively.
We can observe the same tendency as the Ackley function.
Only in Figure~\ref{fig:pgp_holder_table}~(b), the difference between the RMSE of MC estimators is lower than those of other functions.
This is because, since the dimension of the Holder table functions is relatively low, the remaining uncertainty of $f | \*v_t$ is smaller than that in other functions.
Furthermore, we can confirm the computational efficiency of Gibbs sampling also for the Holder table and Hartmann6 functions, as shown in Table~\ref{tab:computation_time_MCMC}.

\begin{table}[!t]
    \caption{The mean and standard deviation of computational time of MCMC for Figure~\ref{fig:pgp_holder_table} (a) and Figure~\ref{fig:pgp_hartmann6} (a) with 10 random trials.}
    \label{tab:computation_time_MCMC}
    \centering
    \begin{tabular} {c|cccc} 
        Time (sec) & LinESS & Gibbs sampling \\ \hline 
        Holder table & $1.71 \pm 0.00$ & $0.55 \pm 0.02$ \\
        Hartmann6 & $1.64 \pm 0.00$ & $0.56 \pm 0.00$
    \end{tabular}
\end{table}

%%%%%%%%%%%%%%%%%%%%%%%%%%%%%%%%%%%%%%%%%%%%%%%%%%%%%%%%%%%%%%%%%%%%%%%%%%%%%%%%%%%%%%%%%%%%%%%%%%%
\subsection{Regret Evaluation for Other Benchmark Functions}

\begin{figure}[t]
    \centering
    \includegraphics[width=0.95\linewidth]{manuscripts/fig_new_compress/Results_legend-min.pdf}

    \subfigure[Computational time (sec)]{
        \includegraphics[width=0.02\linewidth]{manuscripts/fig_new_compress/regret_axis_label-min.pdf}
        \includegraphics[width=0.235\linewidth]{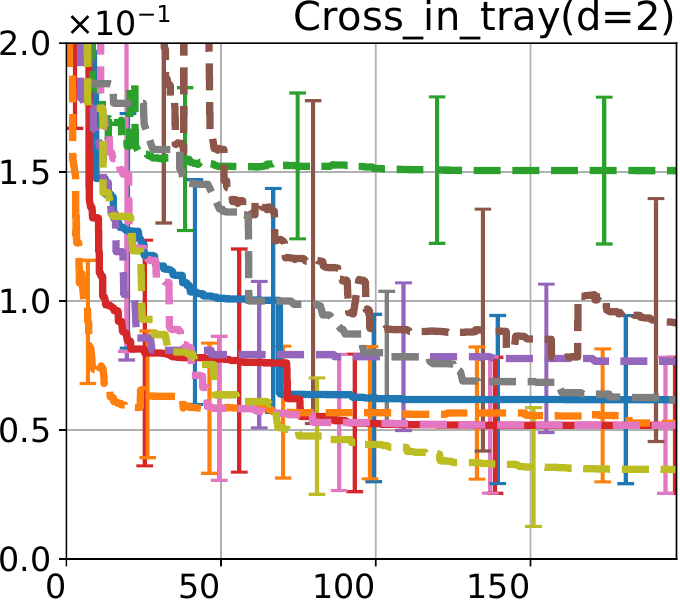}
        \includegraphics[width=0.235\linewidth]{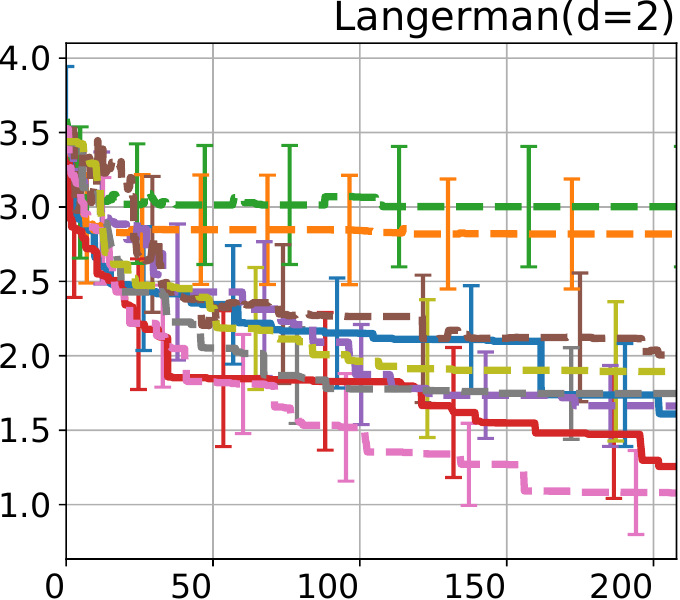}
        \includegraphics[width=0.235\linewidth]{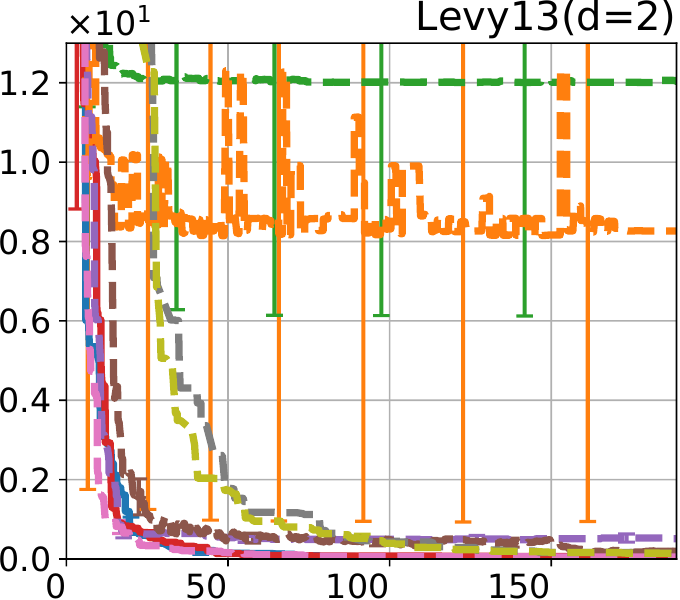}
        \includegraphics[width=0.235\linewidth]{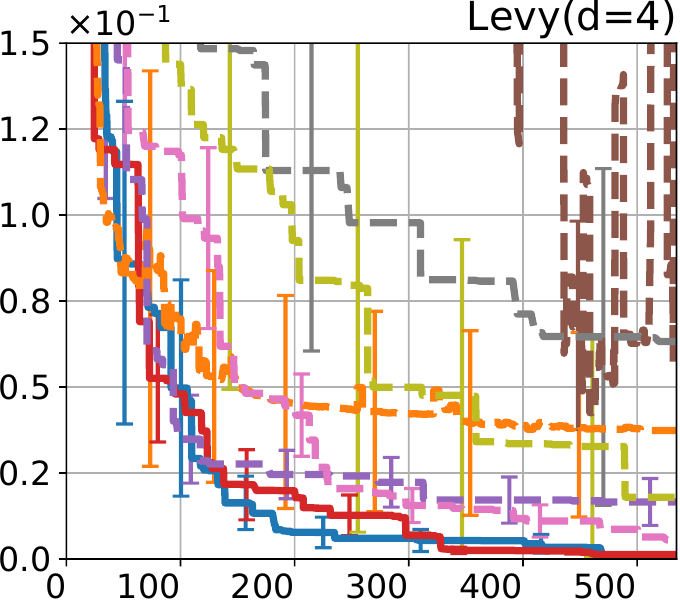}
    }
    \hfill
    \centering
    \subfigure[Iteration]{
        \includegraphics[width=0.02\linewidth]{manuscripts/fig_new_compress/regret_axis_label-min.pdf}
        \includegraphics[width=0.235\linewidth]{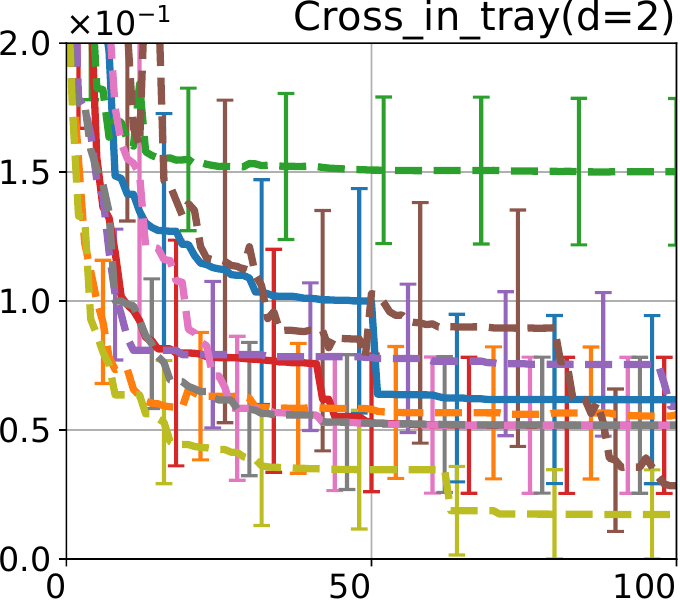}
        \includegraphics[width=0.235\linewidth]{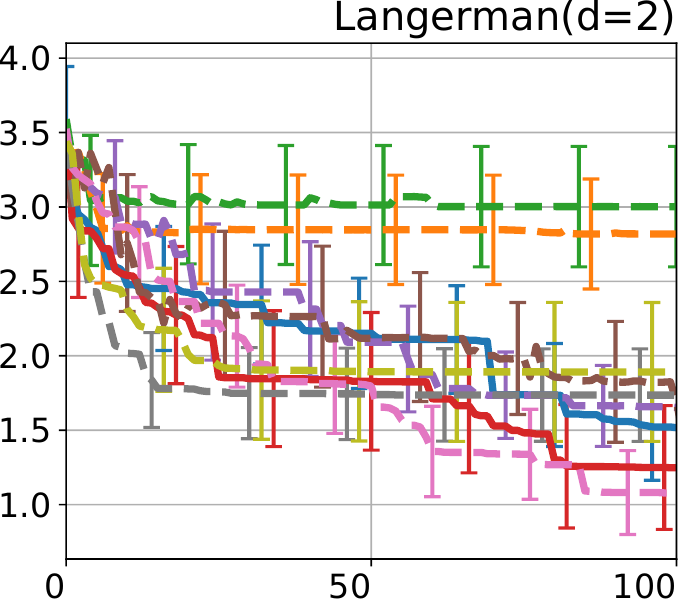}
        \includegraphics[width=0.235\linewidth]{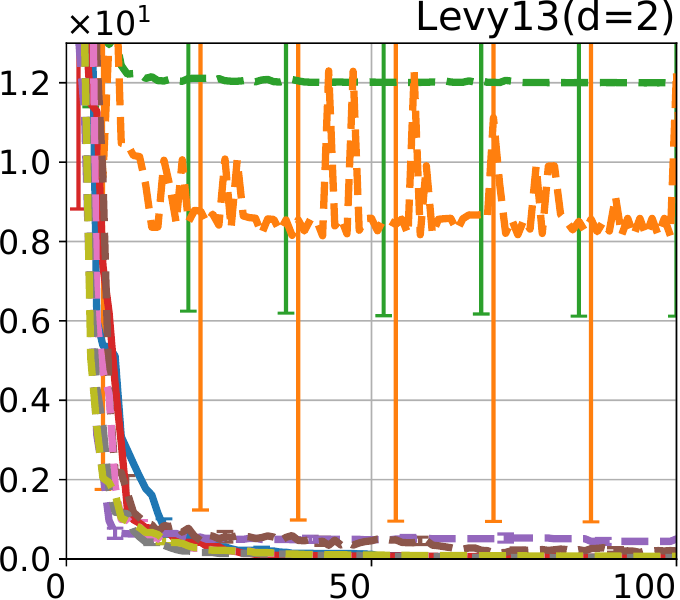}
        \includegraphics[width=0.235\linewidth]{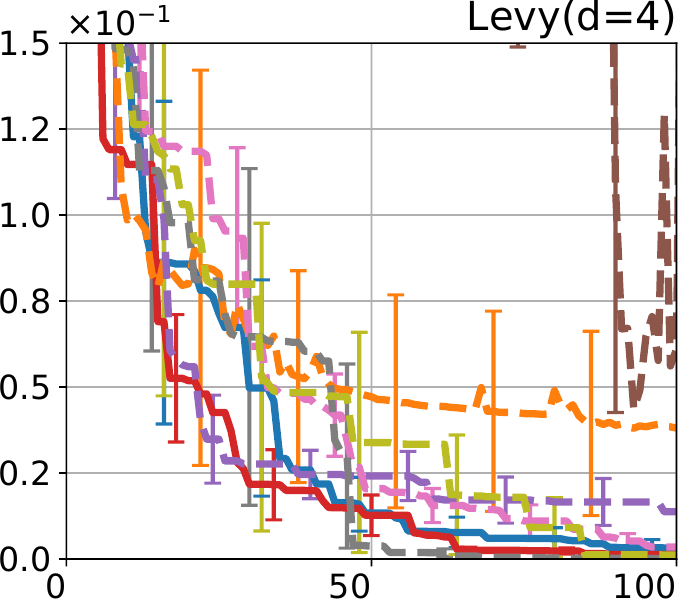}
    }
    
    \caption{Average and standard error of the proposed methods (HB-EI and HB-UCB) and the state-of-the-art preferential BO methods. The horizontal axis represents (a) the computational time (sec) and (b) the number of iterations. 
    The vertical axis represents the regret, which is the smaller, the better it is.}
    \label{fig:app-regret}
\end{figure}

Figure~\ref{fig:app-regret} shows the results for the Cross in tray, Langerman, Levy13, and Levy functions.
In these plots, we can still confirm that the proposed methods, HB-EI and HB-UCB, show superior performance in terms of both computational time and iteration.

%%%%%%%%%%%%%%%%%%%%%%%%%%%%%%%%%%%%%%%%%%%%%%%%%%%%%%%%%%%%%%%%
\subsection{Comparison of HB with Other AFs}
\label{app:comparison_AF}
\begin{figure*}[!th]
    \centering
    \includegraphics[width=0.7\linewidth]{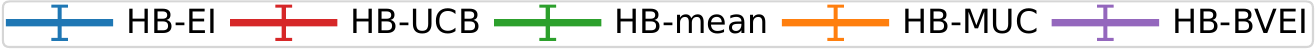}

    \centering
    \includegraphics[width=0.02\linewidth]{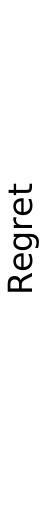}
    \includegraphics[width=0.235\linewidth]{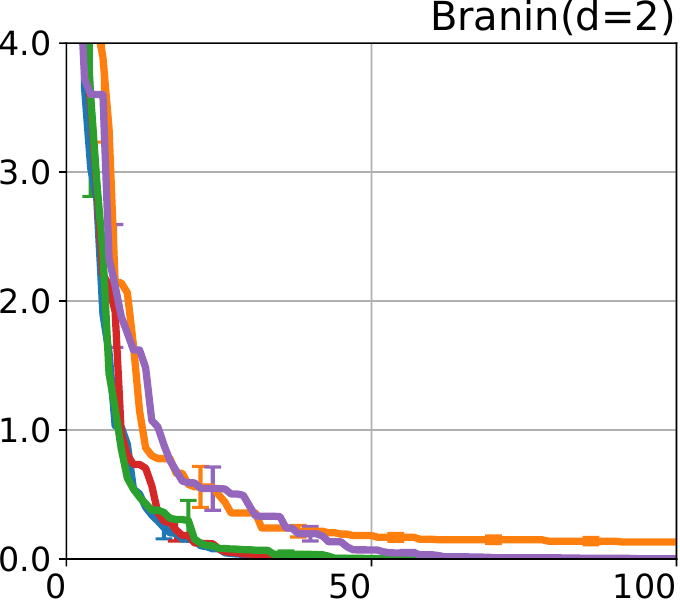}
    \includegraphics[width=0.235\linewidth]{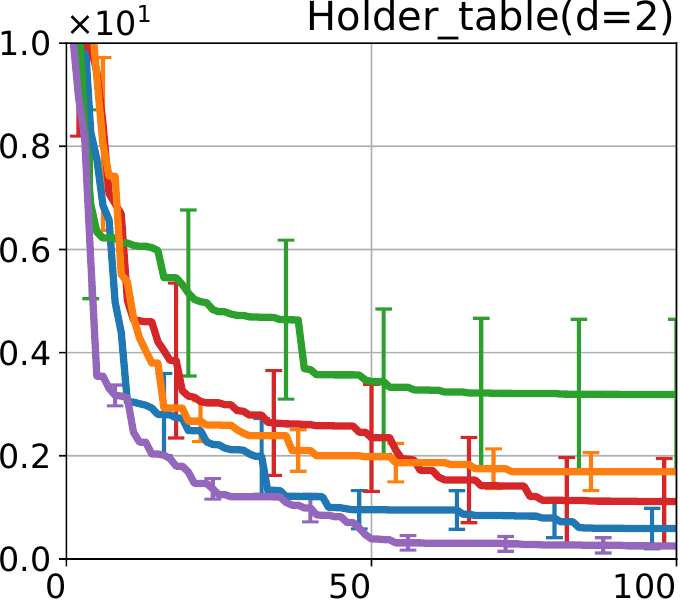}
    \includegraphics[width=0.235\linewidth]{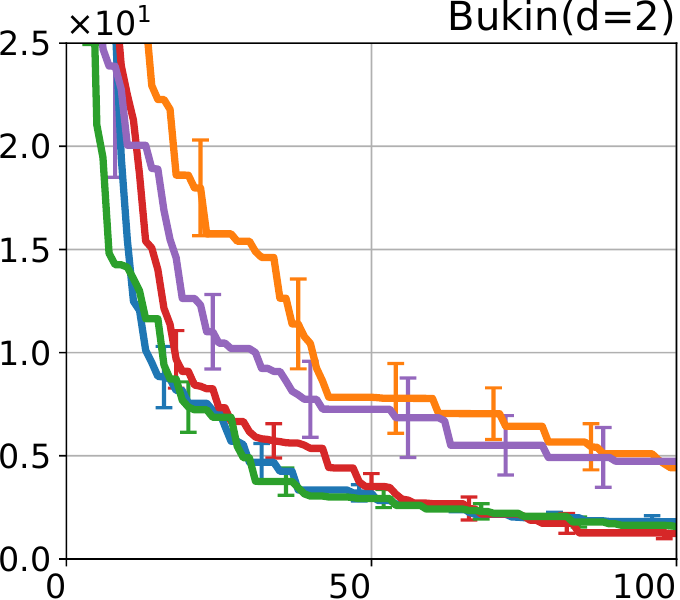}
    \includegraphics[width=0.235\linewidth]{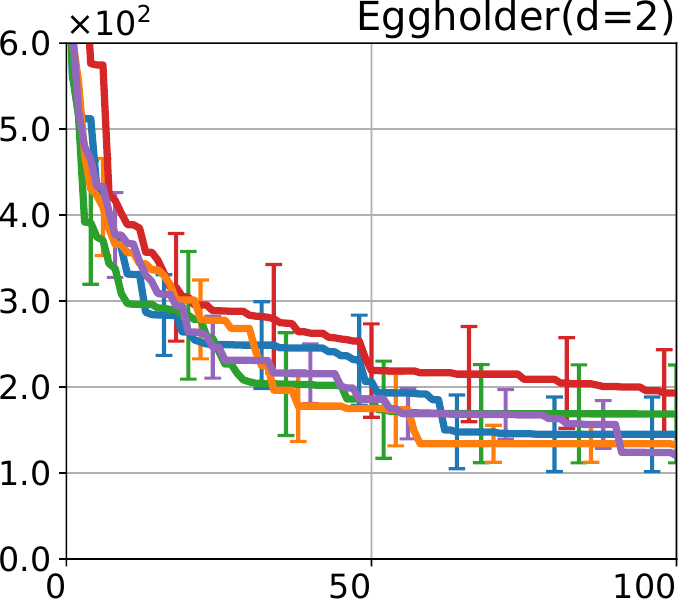}
    \hfill
    \includegraphics[width=0.02\linewidth]{manuscripts/HB_AF_comparison_fig_compress/regret_axis_label-min.pdf}
    \includegraphics[width=0.235\linewidth]{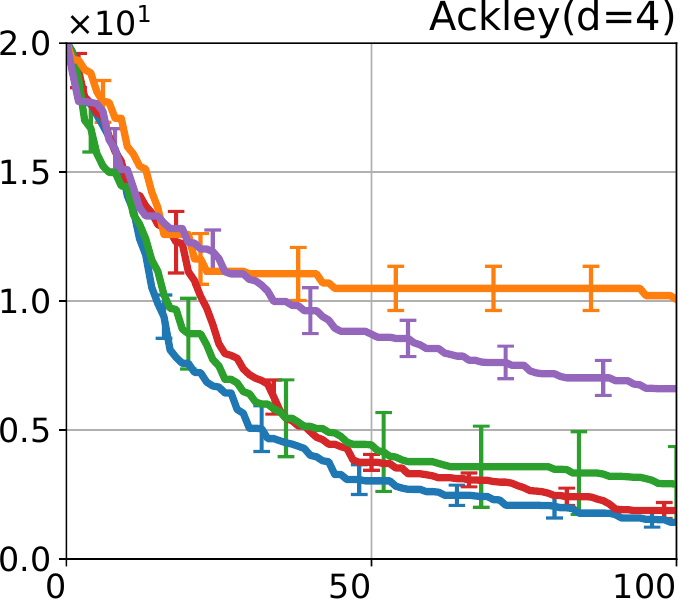}
    \includegraphics[width=0.235\linewidth]{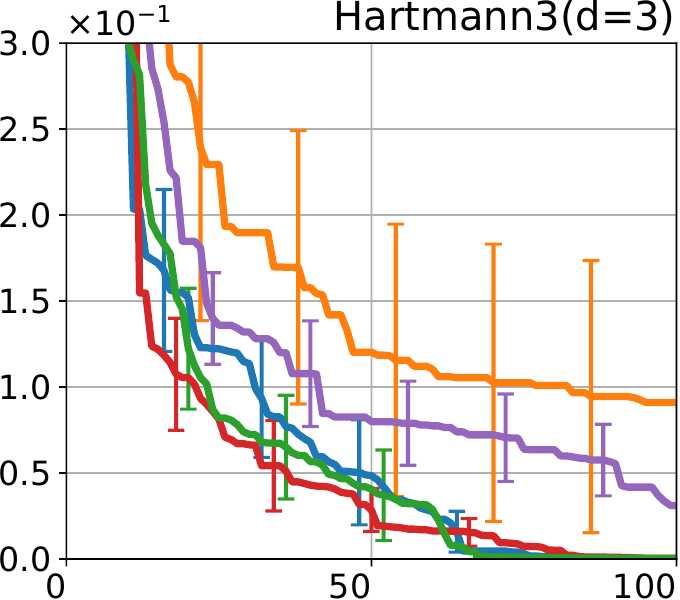}
    \includegraphics[width=0.235\linewidth]{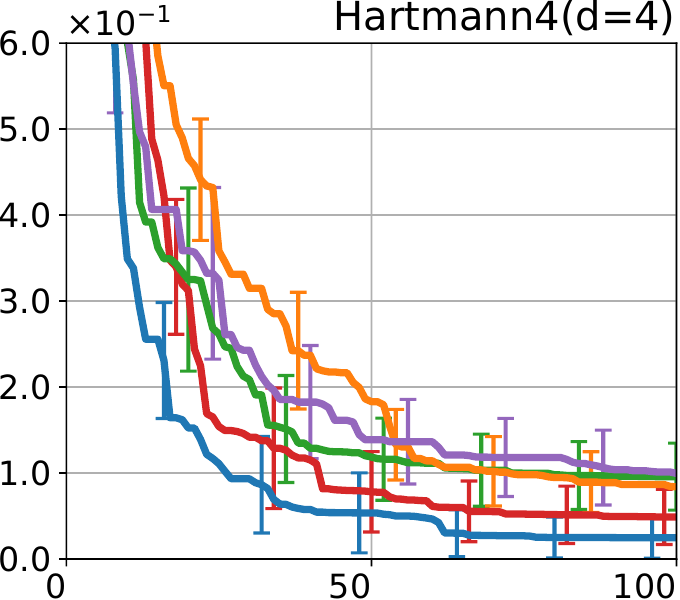}
    \includegraphics[width=0.235\linewidth]{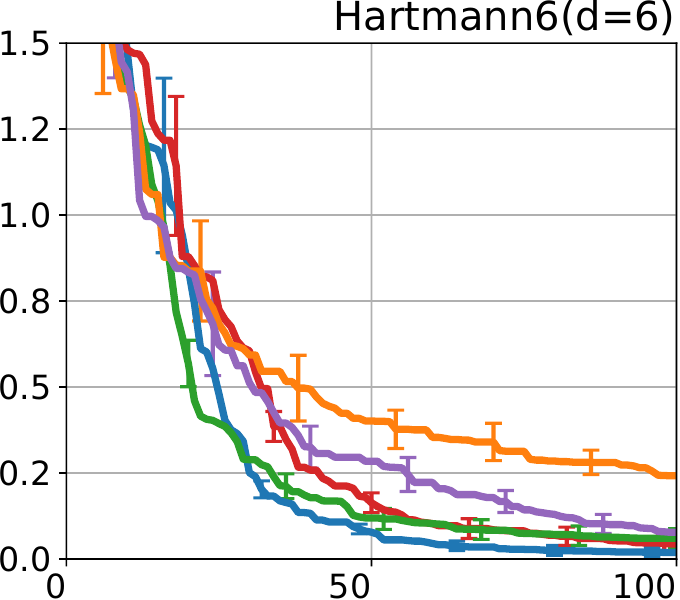}
    \hfill

    \caption{
    Average and standard error of the regret of the proposed HB with several AFs. 
    The horizontal axis represents the number of iterations. 
    The vertical axis represents the regret, which is the smaller, the better it is.}
    \label{fig:regret_HB_AF_comparison}
\end{figure*}

In this section, we evaluate which AF is suitable to combine with HB.
Figure~\ref{fig:regret_HB_AF_comparison} shows the comparisons of HB with several AFs.
Note that, in Figure~\ref{fig:regret_HB_AF_comparison}, the horizontal axis is the iterations since the computational time of each method does not differ largely from each other.
HB-EI and HB-UCB are the same as the methods shown in the main paper and are plotted for comparison.
HB-mean chooses the second input as the maximizer of the mean of $f | \*v_t = \tilde{\*v}_t$.
HB-MUC and HB-BVEI choose the second input using MUC\citep{gonzalez2017preferential,fauvel2021-efficient} and bivariate EI\citep{nielsen2015-perception}, respectively.
All experimental settings are the same as the experiments in Section~\ref{sec:exp}.

We can see that HB-MUC and HB-BVEI are relatively inferior to HB-EI and HB-UCB, particularly in Ackley and Hartmann3 functions.
We conjecture that HB-MUC and HB-BVEI deteriorate by over-exploration, which is caused by the randomization of HB and considering the correlation between the first and second inputs by MUC and BVEI.
On the other hand, HB-mean shows comparable performance with HB-EI and HB-UCB except for the Holder table function.
This could be attributed to HB facilitating the appropriate exploration while the mean maximization is exploitative.

The above experimental results suggest the discussion about the exploit-exploration tradeoff.
Since HB facilitates the exploration, a good choice of the exploit-exploration tradeoff can differ from the usual BO.
A more systematic way to choose the exploit-exploration tradeoff in the HB method is one of the important future works.

